\pdfoutput=1

\documentclass[11pt]{article}

\usepackage[final]{acl}

\usepackage{times}
\usepackage{latexsym}

\usepackage[utf8]{inputenc}

\usepackage{microtype}

\usepackage{inconsolata}

\usepackage{graphicx}
\usepackage{hyperref}       
\usepackage{url}            
\usepackage{booktabs}       
\usepackage{amsfonts}       
\usepackage{nicefrac}       
\usepackage{xcolor}         
\usepackage{comment}

\usepackage{algorithm}
\usepackage{algorithmicx}
\usepackage[noend]{algpseudocode}
\algrenewcommand\alglinenumber[1]{\tiny #1:}
\usepackage{fp}
\usepackage{wrapfig,lipsum,epstopdf}
\usepackage{lipsum}
\usepackage{rotating}
\usepackage{multirow}
\usepackage{pifont}
\usepackage{svg}

\newcommand{\cmark}{\textcolor[rgb]{0.004, 0.58, 0.122}{\ding{51}}} 
\newcommand{\xmark}{\textcolor{red}{\ding{55}}} 
\usepackage{array, makecell} 
\usepackage{xcolor,colortbl}
\usepackage{comment}
\usepackage{kantlipsum}
\usepackage{tabularx}
\usepackage{amsmath}
\usepackage{diagbox}
\usepackage{float}
\usepackage{caption}
\usepackage{amssymb}
\usepackage[most]{tcolorbox}
\usepackage{subcaption}
\usepackage{etoolbox}
\usepackage{enumitem}
\usepackage{adjustbox}
\usepackage{arydshln} 
\setlist{topsep=0pt,noitemsep,topsep=0pt,parsep=0pt,partopsep=0pt}
\definecolor{bgreen}{rgb}{0.0, 0.5, 0.0}
\definecolor{deepskyblue}{rgb}{0.0, 0.75, 1.0}

\newcommand{\papername}{A Benchmark for Large Multi-Modal Models in Long-Form Movies and TV Shows
}
\newcommand{\papernameAbbrev}{InfiniBench}
\newcommand{\infinibenchicon}[1]{\includegraphics[height=35pt]{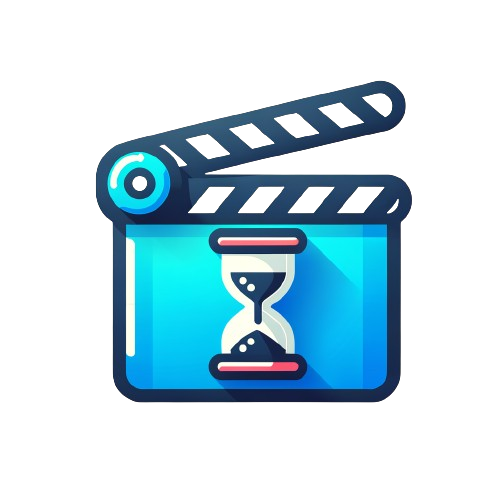}}

\newcommand{\totalSampleNumber}{87.7K}

\title{\infinibenchicon{}{\papernameAbbrev}: {\papername} 
}

\author{
 \textbf{Kirolos Ataallah\textsuperscript{*}\textsuperscript{1}},
 \textbf{Eslam Abdelrahman\textsuperscript{*}\textsuperscript{1}},
 \textbf{Mahmoud Ahmed\textsuperscript{1}},
 \textbf{Chenhui Gou\textsuperscript{2}},
 \\
 \textbf{Khushbu Pahwa\textsuperscript{3}},
 \textbf{Jian Ding\textsuperscript{1}},
 \textbf{Mohamed Elhoseiny\textsuperscript{1}}
\\
\textsuperscript{*}Equal contribution \\
 \textsuperscript{1}KAUST \quad
  \textsuperscript{2}Monash University \quad
 \textsuperscript{3}RICE University \quad
\\ 
}

\begin{document}
\makeatletter
    \let\@oldmaketitle\@maketitle
    \renewcommand{\@maketitle}{\@oldmaketitle
        \myfigure\bigskip}
    \makeatother
    \newcommand\myfigure{%
    \centering
      \includegraphics[width=0.90\linewidth]{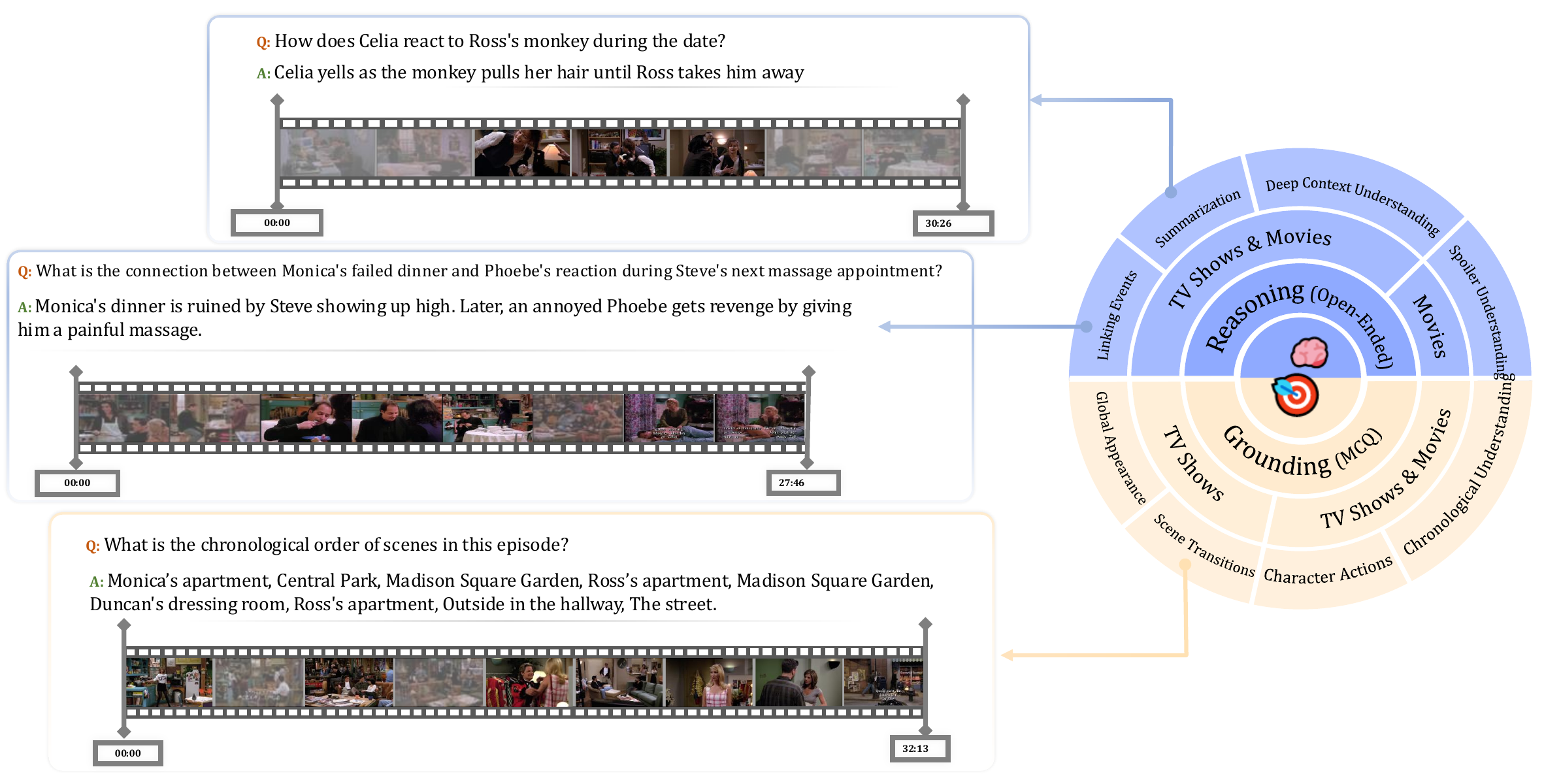}
         \\
          \centering
          \refstepcounter{figure}\normalfont{Figure~\thefigure: {Overview of {\papernameAbbrev} skills: covers eight core skills, grouped into grounding-based (MCQ) and reasoning-based (open-ended) categories.}
          }
        \label{fig:teaser_figure}
}

\maketitle
\begin{abstract}
Understanding long-form videos, such as movies and TV episodes ranging from tens of minutes to two hours, remains a significant challenge for multi-modal models. 
Existing benchmarks often fail to test the full range of cognitive skills needed to process these temporally rich and narratively complex inputs. 
Therefore, we introduce {\papernameAbbrev}, a comprehensive benchmark designed to evaluate the capabilities of models in long video understanding rigorously.
{\papernameAbbrev} offers:
\textit{\textbf{(1) Over 1,000 hours of video content}}, with an average video length of 53 minutes.
\textit{\textbf{(2) The largest set of question-answer pairs}} for long video comprehension, totaling around \totalSampleNumber.
\textit{\textbf{(3) Eight diverse skills}} that span both grounding-based (e.g., scene transitions, character actions) and reasoning-based (e.g., deep context understanding, multi-event linking).
\textit{\textbf{(4) Rich annotation formats}}, including both multiple-choice and open-ended questions.
We conducted an in-depth evaluation across both commercial (GPT-4o, Gemini 2.0 Flash) and most recent open-source vision-language models such as Qwen2.5-VL, InternVL3.0). 
Results reveal that:
(1) Models struggle across the board: Even the best model, GPT-4o, achieves only 47.1\% on grounding-based skills, with most models performing near or just above random chance.
(2) Strong reliance on world knowledge: Models achieve surprisingly high scores using only metadata (e.g., video titles), highlighting a tendency to rely on pre-trained knowledge rather than actual visual or temporal understanding.
(3) Multi-Modal Importance: When provided with full video and subtitle context, however, models show substantial improvements, confirming the critical role of multimodal input in video understanding.

Our findings underscore the inherent challenges in long-video comprehension and point to the need for substantial advancements in both grounding and reasoning capabilities in MLLMs. 
{\papernameAbbrev} is publicly available at \href{https://vision-cair.github.io/Infinibench/}{{\papernameAbbrev}}.
\end{abstract}

\begingroup
\renewcommand{\thefootnote}{}
\footnotetext{Khushbu Pahwa contributed to this work during her internship at KAUST.}
\addtocounter{footnote}{-1}
\endgroup

\begin{table*}[ht!]
    \centering
    \resizebox{0.99\textwidth}{!}{
    \begin{tabular}{ccccccccccccc}
    \toprule
    \multirow{2}{*}{\textbf{Category}} & \multirow{2}{*}{\textbf{Benchmarks}} & \multirow{2}{*}{\textbf{\# Questions}} & \multirow{2}{*}{\textbf{\# Videos}} & \multirow{2}{*}{\makecell{\textbf{Avg Video} \\ \textbf{Duration (mins)}}} & \multirow{2}{*}{\makecell{\textbf{\# Hours}}} & \multicolumn{2}{c}{\textbf{Questions Type}}  & \multicolumn{3}{c}{\textbf{QA Source}} & \multicolumn{2}{c}{\textbf{Annotations}} \\
     &  &  &  &  &  & \textbf{MCQ} & \textbf{Open} & \textbf{Video} & \textbf{Transcript} & \textbf{Summary} & \textbf{Auto} & \textbf{Human} \\
    \cmidrule(r){1-1}
    \cmidrule(r){2-2}
    \cmidrule(r){3-3}
    \cmidrule(r){4-4}
    \cmidrule(r){5-5}
    \cmidrule(r){6-6}
    \cmidrule(r){7-8}
    \cmidrule(r){9-11}
    \cmidrule(r){12-13}
    \multirow{4}{*}{Short} & TGIF-QA \citep{jang2017tgifqa} & 165.2 K & \textbf{72.0 K} & 0.05 & 60.00& \multicolumn{1}{c}{\xmark} & \cmark & \multicolumn{1}{c}{\cmark} & \multicolumn{1}{c}{\xmark} & \xmark & \multicolumn{1}{c}{\cmark} & \cmark \\
    
    \- & MSRVTT-QA \citep{xu2017video}& \textbf{243.6 K} & 10.0 K & 0.25  & 416.6& \multicolumn{1}{c}{\xmark} & \cmark & \multicolumn{1}{c}{\cmark} & \multicolumn{1}{c}{\xmark} & \xmark &\multicolumn{1}{c}{\cmark} & \xmark \\
    
    \- & MV-Bench \citep{li2024mvbench} & 4.0 K & 3.6 K & 0.27 & 16.38& \multicolumn{1}{c}{\cmark} & \xmark & \multicolumn{1}{c}{\cmark} & \multicolumn{1}{c}{\xmark} & \xmark & \multicolumn{1}{c}{\cmark} & \xmark \\ 
    \- & TVQA \citep{lei2019tvqa} & 152.5 K & 21.8K & 1.27  & 461.20 & \multicolumn{1}{c}{\cmark} & \xmark & \multicolumn{1}{c}{\cmark} & \multicolumn{1}{c}{\xmark} & \xmark & \multicolumn{1}{c}{\xmark} & \cmark \\
    
    \midrule
    \multirow{8}{*}{Long} & Activity-QA \citep{yu2019activitynetqa}& 58.0 K & 5800  & 3.00 & 290.00& \multicolumn{1}{c}{\xmark} & \cmark & \multicolumn{1}{c}{\cmark} & \multicolumn{1}{c}{\xmark} & \xmark & \multicolumn{1}{c}{\xmark} & \cmark \\

    \- & Egoschema \citep{mangalam2023egoschema}& 5.0 K & 5063 & 3.00 &253.15& \multicolumn{1}{c}{\cmark} & \xmark & \multicolumn{1}{c}{\cmark} & \multicolumn{1}{c}{\xmark} & \xmark & \multicolumn{1}{c}{\cmark} & \cmark \\
    
    \- & LongVideoBench \citep{wu2024longvideobench}& 6.7 K & 3763 & 7.88 & 494.21& \multicolumn{1}{c}{\cmark} & \xmark & \multicolumn{1}{c}{\cmark} & \multicolumn{1}{c}{\xmark} & \xmark & \multicolumn{1}{c}{\xmark} & \cmark \\
    
    \- & Moviechat \citep{song2023moviechat} & 14.0 K & 1000 & 9.40  &156.67& \multicolumn{1}{c}{\xmark} & \cmark & \multicolumn{1}{c}{\cmark} & \multicolumn{1}{c}{\xmark} & \xmark & \multicolumn{1}{c}{\xmark} & \cmark \\
    
    \- & MLVU \citep{zhou2024mlvucomprehensivebenchmarkmultitask} &3.1 K& 1730&15.50  &446.92& \multicolumn{1}{c}{\cmark} & \cmark & \multicolumn{1}{c}{\cmark} & \multicolumn{1}{c}{\xmark} & \xmark & \multicolumn{1}{c}{\cmark} & \cmark \\ 
    
    \- &MoVQA \citep{zhang2023movqa}&21.9 K&100&16.53 &27.55& \multicolumn{1}{c}{\cmark} & \xmark & \multicolumn{1}{c}{\cmark} & \multicolumn{1}{c}{\xmark} & \xmark & \multicolumn{1}{c}{\xmark} & \cmark \\ 
    
    \- &Video-MME \citep{videomme}&2.7 K& 900 & 16.97 &254.55& \multicolumn{1}{c}{\cmark} & \xmark & \multicolumn{1}{c}{\cmark} & \multicolumn{1}{c}{\xmark} & \xmark & \multicolumn{1}{c}{\xmark} & \cmark \\
    \midrule
    
    Very Long & LVBench \citep{wang2024lvbench}&1.6 K& 103&\textbf{68.35}&  117.33& \multicolumn{1}{c}{\cmark} & \xmark & \multicolumn{1}{c}{\cmark} & \multicolumn{1}{c}{\xmark} & \xmark & \multicolumn{1}{c}{\xmark} & \cmark \\

    \rowcolor{blue!15} \cellcolor{white} &  \textbf{{\papernameAbbrev} (Ours)} & \textbf{\totalSampleNumber} & 1217 & 52.59 &\textbf{1066.70}&\multicolumn{1}{c}{\cmark} & \cmark & \multicolumn{1}{c}{\cmark} & \multicolumn{1}{c}{\cmark} & \cmark & \multicolumn{1}{c}{\cmark} & \cmark \\
    
    \bottomrule
    \end{tabular}
    }
    \caption{Comparison between {\papernameAbbrev} and existing video QA benchmarks and datasets.
    {\papernameAbbrev} contains the largest number of QA pairs for long and very long videos, as well as the longest total video duration.}
    \label{tab:benchmark_comparisons}
    \vspace{-3mm}
\end{table*}

\section{Introduction}
Recent progress in Multimodal large language models (MLMMs) has enabled impressive performance on image and short-video understanding by jointly processing visual and textual inputs~\citep{minigpt4_video,zhu2023minigpt,video_llama,minigptv2,video-llava,llava,maaz2023video}. To push toward richer video comprehension, several models have extended the context length of underlying LLMs, enabling them to process longer videos ~\citep{bai2023qwenvl,bai2025qwen25vltechnicalreport,li2024llavaonevisioneasyvisualtask}. However, current public benchmarks remain limited in both scope and realism. Most are restricted to short clips (a few seconds to minutes), and even larger-scale efforts like LVBench, with one-hour videos, lack narrative complexity and modality diversity~\citep{wang2024lvbench}.
In real-world scenarios, long-form videos especially movies and TV shows present fundamentally different challenges. These formats feature rich, structured storytelling with long-range temporal dependencies, evolving character arcs, and intricate causal relationships~\citep{zhang2023movqa,lei2019tvqa,song2023moviechat}
. Unlike short videos, they demand high-level skills such as intent inference, multi-event linking, and thematic summarization. Humans can easily integrate dispersed cues across modalities and time; current MLMMs cannot.

To bridge this gap, we introduce {\papernameAbbrev}, a large-scale benchmark purpose-built to evaluate the long-video understanding capabilities of multi-modal models. {\papernameAbbrev} leverages over 1,000 hours of content drawn from movies and TV series with average video length (52.59 minutes) and the largest set of skill-targeted question–answer pairs (\totalSampleNumber) among existing long and very long video benchmarks (see Table~\ref{tab:benchmark_comparisons}).

Crucially, {\papernameAbbrev} is organized around eight core skills necessary for holistic long-video understanding. These are grouped into two categories:
(1) Grounding-based skills, which assess a model’s ability to retrieve and organize visual and temporal information, and
(2) Reasoning-based skills, which require causal inference, contextual understanding, and abstract reasoning.

Key Insights from {\papernameAbbrev}:
Despite the scale and capabilities of modern MLMMs, all evaluated models struggle on {\papernameAbbrev}. 
For instance, GPT-4o, which is the top-performing commercial model, achieves only 47.1\% accuracy on grounding-based tasks, just modestly above the random baseline of 20\%. This performance gap highlights how far even the best models are from mastering long-video comprehension.
To dissect model behavior further, we conduct controlled experiments that isolate the role of world knowledge vs. visual input. Surprisingly, models still achieve non-trivial accuracy when given only metadata (e.g., show title and episode number), indicating heavy reliance on pre-trained knowledge. 
Yet, the most substantial gains up to +15\% in accuracy come from providing visual input, confirming that true video understanding requires visual grounding, not just memorized associations.
Finally, we observe consistent underperformance across both skill groups, grounding and reasoning-based skills. 
Reasoning-based tasks remain especially challenging, with even commercial models failing to produce coherent, causal, or context-aware answers. These findings underscore the unique difficulty and diagnostic power of {\papernameAbbrev}, which surfaces fundamental limitations in today’s multi-modal models and points toward critical future directions.

\section{Related Work}


In this section, we cover the existing video benchmarks and position our work among them.
An extended version of the related work can be found in Section \ref{sec:appendix_relatedwork}.

Recently, many multimodal large language models have aimed to extend their context lengths, such as Qwen2.5-VL~\citep{bai2025qwen25vltechnicalreport}, Qwen2-VL~\citep{wang2024qwen2vlenhancingvisionlanguagemodels}, and LongVU~\citep{wu2024longvideobench}, which can process videos over an hour long. This highlights the growing need for effective benchmarks. In contrast, short video benchmarks have been widely explored, covering tasks like moment retrieval~\citep{liang2024tvrrankingdatasetrankedvideo,lei2021qvhighlightsdetectingmomentshighlights}, action classification~\citep{caba2015activitynet,kinetics-400,soomro2012ucf101dataset101human}, video reasoning~\citep{xie2024funqasurprisingvideocomprehension,xiao2021nextqanextphasequestionansweringexplaining}, and video QA~\citep{xu2017video,lei2019tvqa,jang2017tgifqa}. Among longer video benchmarks, TVQA~\citep{lei2019tvqa} is an early example with over 152.5 K QA pairs across 21.8 K clips (1.27 min avg.), totaling 461.2 hours. More recent datasets like MovieChat-1K\citep{song2023moviechat} offer 1,000 movie clips (9.4 min avg.) with 14K annotations. CRAFT\citep{ates2020craft} and GazeVQA~\citep{ilaslan2023gazevqa} extend to causal and embodied reasoning. Other long-form efforts include MLVU~\citep{zhou2024mlvucomprehensivebenchmarkmultitask}, MoVQA~\citep{zhang2023movqa}, and Video-MME~\citep{videomme}, though their average durations peak at 17 minutes. LVBench\citep{wang2024lvbench} stands out with 1-hour videos, but includes only 103 clips totaling 117 hours. Domain-specific datasets like Ego4D QA on EgoSchema\citep{mangalam2023egoschema} focus on first-person views but use short (3-minute) clips. Motivated by these gaps, our benchmark evaluates grounding (temporal/visual anchoring) and reasoning (causality, narrative) skills to reflect human-like video understanding. As shown in Table~\ref{tab:benchmark_comparisons}, our dataset supports near-hour-long videos and offers the most significant total duration ($\sim$1K hours) and the largest QA-pair count in terms of long video understanding (\totalSampleNumber).

    \section{{\papernameAbbrev}}

\subsection{Essential Skills for Long-Video Understanding}
\label{sec_method_skills_def}
Understanding long videos requires more than event recognition, it demands tracking narratives, entities, and interactions over time.
We categorize the core skills into two groups:
1- Grounding-Based Skills, focused on retrieving and organizing information directly from the video; and
2- Reasoning-Based Skills, which require inference, contextual understanding, and causal analysis.
This division mirrors human comprehension: grounding what happens, then reasoning about why. {\papernameAbbrev} leverages this structure to offer a comprehensive evaluation of long-video understanding.

\noindent\subsubsection{Grounding-Based Skills}
Grounding-based skills evaluate a model’s ability to retrieve, order, and structure video content without requiring inference. They reflect the foundation of narrative comprehension, recognizing what happened, in what order, and to whom.

\noindent\textbf{Chronological Understanding:}
Measures a model’s ability to sequence events. This mirrors a core human faculty: understanding stories requires ordering moments in time. Without temporal grounding, narrative comprehension breaks down.

\noindent\textbf{Character Actions Tracking:}
Assess whether the model can group and order actions performed by a specific character. Understanding ``who did what, and when'' is key to following character arcs and maintaining narrative coherence.

\noindent\textbf{Scene Transitions:}
Focuses on recognizing and organizing scene-level shifts. Accurate modeling of transitions enables a higher-level understanding of narrative flow, pacing, and structural progression across long videos.

\noindent\textbf{Global Appearance:}
Tests the ability of the model to track changes in character appearance throughout the video. This requires long-term visual consistency and awareness of visual storytelling cues, e.g., ``What is the change in outfit for (character-name) in this video?".

\subsubsection{Reasoning-Based Skills}
These skills assess deeper narrative understanding, where models must infer relationships, motivations, and implications not explicitly stated. They reflect how humans interpret stories by connecting context, prior knowledge, and unstated cues.

\noindent\textbf{Linking Multiple Events:}
Evaluates the ability to connect distant or seemingly unrelated events by inferring causality.
Example: ``How does Chandler's previous experience with his own parents' separation influence his decision to break up with Janice?"
A: ``Chandler’s experience with his parents' separation strongly influences his decision. Remembering how he hated the man who came between his parents, Chandler decides to step aside for the sake of Janice's family's happiness, which leads him to break up with her."

\noindent\textbf{Deep Context Understanding:}
Tests whether a model can interpret hidden motivations and social cues.
Example: Q: How did Ted's co-workers' attitudes towards him change after he became their boss? A: Ted's co-workers clammed up whenever he walked into the room and became distant judging him on everything he did or said.

\noindent\textbf{Spoiler Questions:}
Determines if a model can recognize when a question reveals a major plot twist or ending.
Example: Q: Why did Roth betray Michael? A: It is implied that Roth wanted revenge for the death of Moe Greene, a friend of his who was assassinated on Michael's orders.

\noindent\textbf{Summarization:}
Assesses the model’s ability to distill narratives into summaries that capture events, emotional arcs, and thematic depth. Unlike short clips, movies and episodes demand multi-layered abstraction, not just factual recall.

For more visual samples for each skill, see \ref{sec:appendix_skills_examples} in the supplementary.
 \begin{figure}[t!]
    \centering
    \includegraphics[width=1\linewidth]{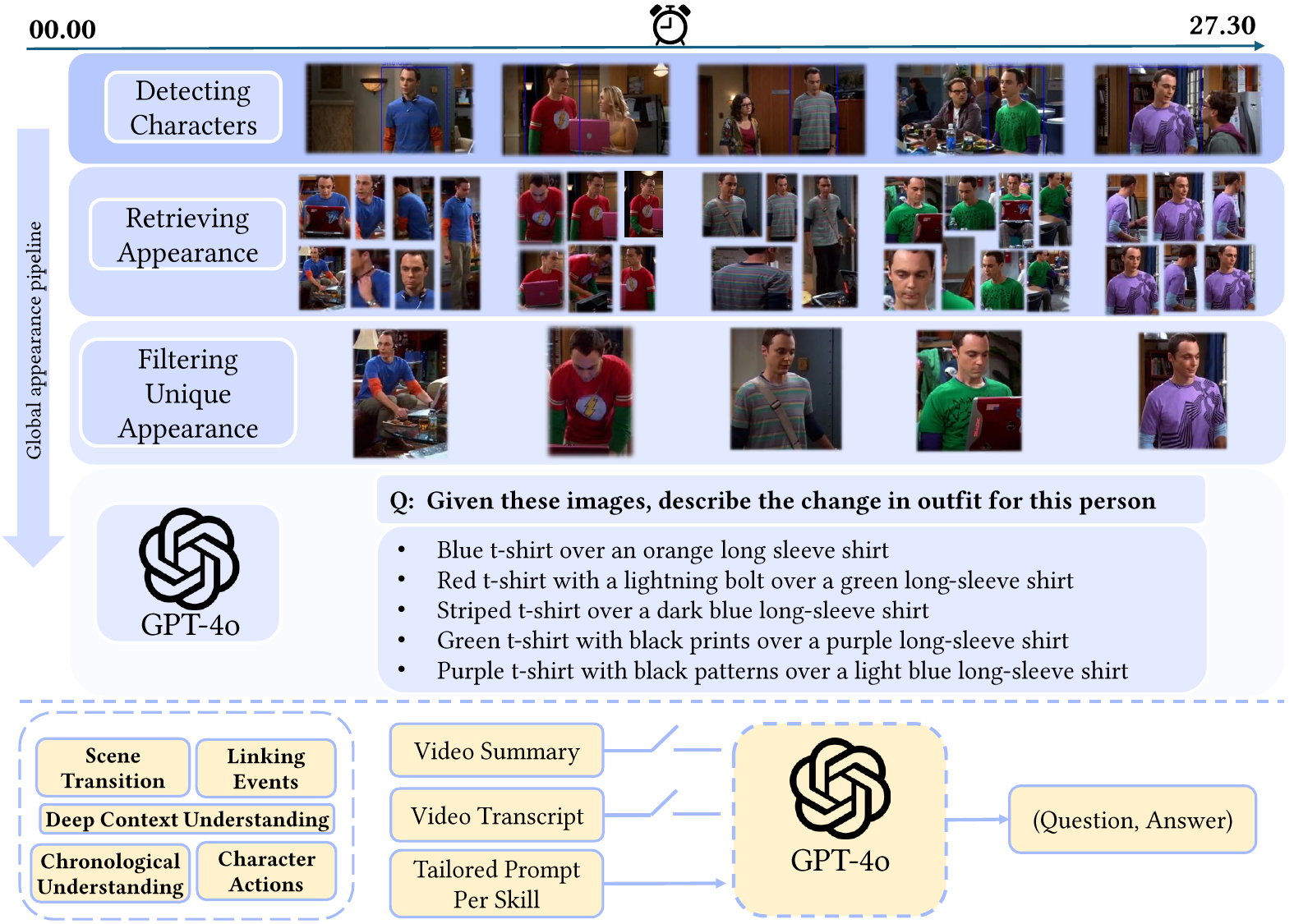}
    \caption{Full annotation pipeline for {\papernameAbbrev} skill set. The upper section depicts the global appearance pipeline, while the lower section illustrates the question generation using GPT-4o. The gates for video summary and video transcript indicate that some skills utilize only the summary, others use only the transcript, and some use both.}
    \label{fig:annotation_pipeline}
    \vspace{-4mm}
\end{figure}

\subsection{Data Collection Pipeline}
\label{sec_method_data_collection}
Building a rigorous long-video benchmark demands data that mirrors real-world storytelling, with multi-character dynamics, evolving plots, and implicit causal links. 
Our pipeline ensures diversity, depth, and relevance through three steps:
1- Video Selection (Sec.~\ref{sec_method_video_sources}): Choose movies and TV episodes that exhibit complex narratives.
2- QA Generation (Sec.~\ref{sec_method_qa_generation}): Craft question–answer pairs aligned with each of the eight skills.
3- Refinement (Sec.~\ref{sec_method_data_integrity}): Remove redundancies, and the questions that can be easily answered without the visual signal, e.g., from the subtitles only or from the model knowledge.

\subsubsection{Why Movies and TV-Shows}
\label{sec_method_video_sources}
Unlike vlogs or surveillance footage, which are often repetitive and low in narrative complexity, movies and TV shows offer the rich, structured storytelling essential for long-video understanding.
They feature:
1- Diverse, Long-Term Contexts: Multi-layered plots and evolving character arcs require memory and integration across time.
2- Non-Linear Storytelling: Flashbacks, subplots, and temporal shifts challenge models to reason over fragmented, distant events.
These properties make them ideal for evaluating both grounding and reasoning skills in extended narratives.


\subsubsection{Generating Question-Answer Pairs}
\label{sec_method_qa_generation}
We construct structured question–answer pairs tailored for long-video understanding, where models must reason over extended contexts, not just recognize isolated moments. Our pipeline, as shown in Figure \ref{fig:annotation_pipeline}, leverages the video visual-signal combined with transcripts and summaries.

\noindent\textbf{Why Transcripts?}
Transcripts offer richer context than subtitles, including scene descriptions and character actions (e.g., ``Monica and Chandler enter the café''). 
These cues allow for generating deeper, temporally grounded questions that reflect both visual and structural aspects of the story. For more details, see \ref{sec:diff_trans_sub} in the supplementary.
For the TV shows, transcripts were sourced from reputable platforms such as \cite{foreverdreaming} and manually verified and aligned with the video.
For movies, we utilize the transcripts provided by MovieNet \cite{huang2020movienet}, which have been pre-processed to ensure accurate alignment with both dialogue and subtitles, as detailed in Section A.5 \cite{huang2020movienet}.

\noindent\textbf{Chronological Understanding \& Linking Events:}
We employ GPT-4o to chronologically extract key events from the transcript. 
Then, we generate multiple-choice questions that assess the understanding of the full or partial chronological sequence. For linking events, we input GPT-4o the extracted events and asked the model to connect non-adjacent events by identifying causal relations and generate the Q/A pairs.

\begin{figure*}[htbp]
    \centering
    \captionsetup{font=small}  
    \begin{minipage}{0.74\textwidth}
        \centering
        \includegraphics[width=0.99\textwidth]{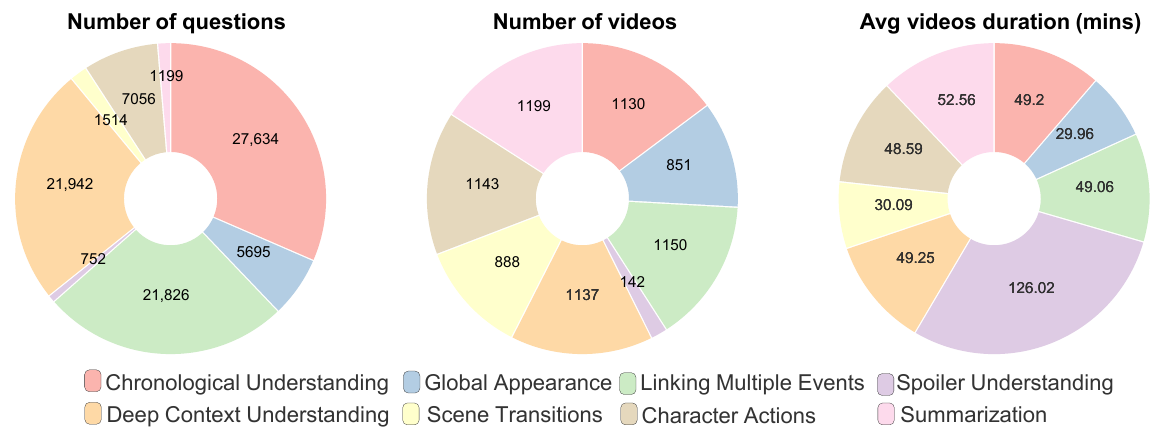}
        \caption{{\papernameAbbrev} skills statistics. (A) Number of questions per skill, (B) Number of videos per skill, and (C) Average video duration per skill}
        \label{fig:skills_new}
    \end{minipage}
    \hfill
    \begin{minipage}{0.24\textwidth}
        \centering
        \includegraphics[width=0.99\textwidth]{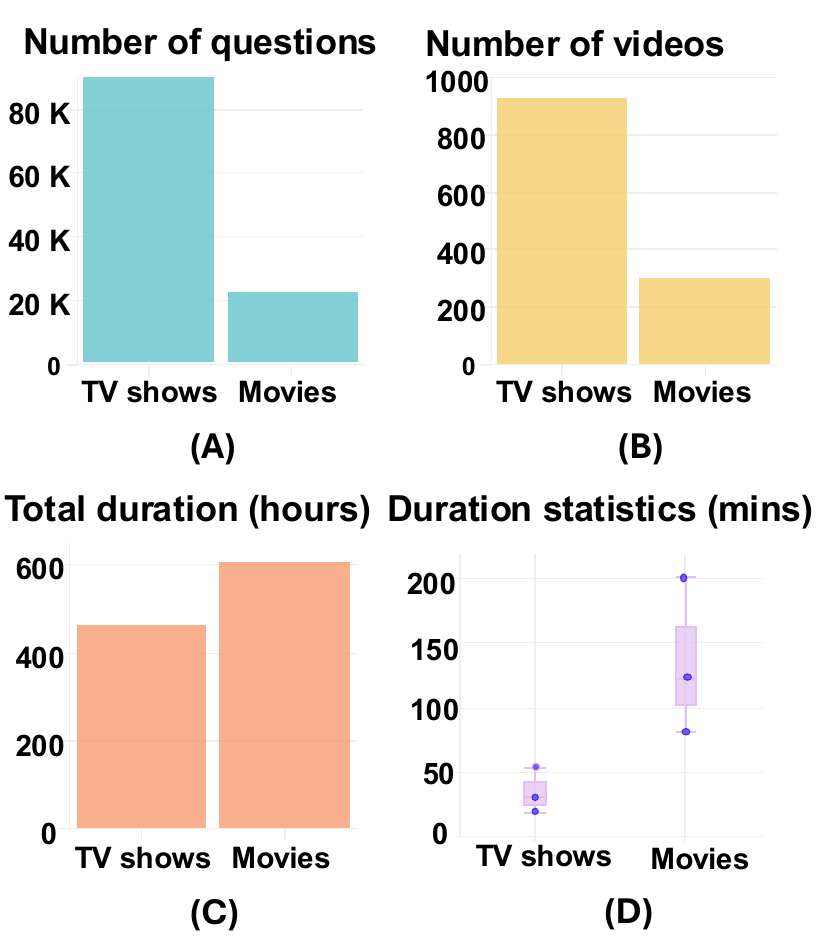}
        \caption{TV shows and Movies statistics.}
        \label{fig:shows_vs_videos}
    \end{minipage}
\end{figure*}

\noindent\textbf{Scene Transitions}:
We prompted GPT-4o to identify all scene locations appearing in the transcript in chronological order. Based on this information, we generated multiple-choice questions designed to assess the correct sequential ordering of these locations in the video.

\noindent\textbf{Character Tracking}:
To track character actions in the video, we input the transcript and the video summary into GPT-4o, which generates question-answer pairs focused on identifying the sequence of actions performed by the main characters.
\noindent\textbf{Deep Context Understanding:}
For the deep context understanding question-answer pairs, we employed GPT-4o with the transcript and the summary to generate questions about the hidden motivations and social cues.
\noindent\textbf{Global Appearance:}
Our pipeline, as shown in Figure \ref{fig:annotation_pipeline}, integrates person detection \cite{khanam2024yolov11overviewkeyarchitectural} and face-matching using InsightFace \cite{guo2021sample} to identify characters in video frames. Reference images of the main cast were collected from IMDb and matched with detected faces, using an empirically determined threshold to filter out unmatched individuals. Cropped character images were saved chronologically in timestamped folders.
To identify unique outfits, we extracted visual features using DINO-v2 \cite{oquab2024dinov2learningrobustvisual} and removed redundant frames based on similarity scores, reducing approximately 200 frames to 20 while conservatively preserving visual diversity. The remaining images were then processed with GPT-4o \cite{GPT-4o} to generate outfit descriptions.
Multiple-choice questions were constructed by varying the sequence of outfits. In cases where a character wore a single outfit throughout, distractor options featuring incorrect outfits were introduced to maintain question complexity.
\noindent\textbf{Summarization \& Spoiler Questions:}
We curate expert-written IMDb \cite{imdb} summaries for human-level summarization. Spoiler questions are sourced from flagged plot discussions, focusing on twists or endings.
\subsubsection{Data Integrity and Diversity}
\label{sec_method_data_integrity}

To ensure the reliability and quality of our benchmark, the collected dataset undergoes several refinement steps aimed at eliminating redundancy, mitigating data leakage, and reinforcing multimodal reasoning.

\noindent\textbf{1) Redundancy Removal.}
To avoid question duplication, we compute pairwise textual similarity using BGE-M3 embeddings \cite{bge-m3} and remove highly similar questions (see Section~\ref{sec_appendix_duplication_filtering} for details).
Additionally, since movies often contain over 100 events, many of which are redundant, we apply the same similarity-based filtering to retain only 20 diverse and representative events per movie.

\noindent\textbf{2) Filtering Internal Model's Knowledge.}
Large video-language models are trained on massive datasets and may memorize content from widely available media, raising the risk of data leakage. 
To address this, we evaluate three powerful models—GPT-4o, Gemini 2, and Qwen2.5-VL—on our dev and test splits using only the movie/TV show metadata (e.g., title, genre, year) as input.
If all three models correctly answer a question using this metadata alone, we consider the question likely answerable from internal knowledge and remove it.
This step filters out approximately 16\% of the questions, resulting in a subset we refer to as the \textit{Knowledge-Filtered Set}.

\noindent\textbf{3) Making Vision Matters.}
To ensure that our benchmark requires visual understanding—beyond textual reasoning, we further filter out questions answerable using subtitles alone. 
Using the same three models, we provide only the subtitle stream and explicitly instruct the models to rely solely on this modality.
As in the previous step, questions correctly answered by all three models are excluded. 
This removes an additional 16\%, leading to a total reduction of 32\% across both filtering stages.
The remaining set constitutes our final benchmark, on which all evaluations and ablations are performed unless otherwise noted.

\noindent4) \textbf{Human Verification.}
To further ensure the validity, clarity, and correctness of the dataset, we conducted manual verification on the test set, covering 20\% of the entire benchmark. This verification process revealed a 90.11\% accuracy rate for valid question–answer pairs across different skills, demonstrating high overall reliability. Importantly, any identified errors during this process were filtered out, ensuring that the final test subset reflects a clean and trustworthy evaluation set. 
Additional details can be found in Section~\ref{sec_appendix_human_verf}. The custom verification interface used in this process is shown in Figure~\ref{fig:website_gui}.


\subsection{Data Statistics}
\label{sec_method_data_statistics}
{\papernameAbbrev} ultimately comprises {\totalSampleNumber}  questions over videos averaging 53 minutes, with some running as long as 201 minutes, making it the largest and most challenging benchmark for long-video understanding.
{\papernameAbbrev} includes 923 episodes from six TV shows \citep{lei2019tvqa} based on the long-form version in GoldFish~\cite{ataallah2024goldfishvisionlanguageunderstandingarbitrarily} and filtered 296 movies that have subtitles from MovieNet \citep{huang2020movienet}.
These movies offer a broad and challenging foundation for benchmarking long-video understanding.
Figures \ref{fig:skills_new} and \ref{fig:shows_vs_videos} illustrate the per-skill distribution, source breakdown (TVQA \citep{lei2019tvqa}, MovieNet \citep{huang2020movienet}), and duration statistics across the dataset.
We humanly verified 20\% of the data, ensuring fair coverage across TV shows and movies. The remaining 80 \% is available for training video models. The verified portion is further divided into test and validation splits, 10\% each. 
All reported results and ablation studies are conducted using the test split.

\section{Experiments}

\subsection{Models}
To assess the capabilities and limitations exposed by {\papernameAbbrev}, we evaluate 13 state-of-the-art long-video MultiModal Large Language Models (MLLMs) spanning both open-source and commercial ecosystems.

\noindent\textbf{Open-Source Models}
We benchmark eleven leading open-source MLLMs, each evaluated using their official implementations and following their official best practices and configuration, e.g., recommending sampling strategy and frame-rate. 
These models vary in architecture, context length, and vision-language alignment strategies, offering a diverse perspective on the current research landscape:
Qwen2.5-VL \cite{bai2025qwen25vltechnicalreport}, Qwen2-VL \cite{wang2024qwen2vlenhancingvisionlanguagemodels}, InternVL3 \cite{zhu2025internvl3}, InternVL2.5 \cite{InternVL2.5}, InternVL2\cite{InternVL2}, LLaVA-OneVision \citep{li2024llavaonevisioneasyvisualtask}, LongVU \citep{shen2024longvuspatiotemporaladaptivecompression},VideoChat-Flash \cite{li2025videochatflashhierarchicalcompressionlongcontext}, InternLM-XComposer-2.5 \cite{zhang2024internlmxcomposer25versatilelargevision}, Goldfish \cite{ataallah2024goldfishvisionlanguageunderstandingarbitrarily},  and MiniGPT4-video\cite{minigpt4_video}

\noindent\textbf{Commercial Models}
We also evaluate two cutting-edge proprietary models via their official APIs:
GPT-4o~\cite{GPT-4o} and 
Gemini 2.0 Flash ~\cite{gemini_flash}.
These models represent the frontier of closed-access MLLMs with advanced capabilities in video, text, and multi-turn interaction. Their inclusion offers a valuable benchmark ceiling for current open-source efforts.

We followed the standard best practices for all the models based on their official implementation.
More details can be seen in Section \ref{sec:models_config}.
\subsection{Evaluation Setup}
We adopt distinct evaluation protocols for grounding-based and reasoning-based skills, reflecting their differing task formats: multiple-choice (MCQ) and open-ended, respectively.

\noindent\textbf{Multiple-Choice (MCQ).}  
For grounding-based skills, model responses are evaluated using standard accuracy. Models are prompted to output the correct option number, and accuracy is computed based on an exact match with the correct answer.

\noindent\textbf{Open-Ended.}
Reasoning-based skills are assessed using a GPT-4o mini, assigning a score from 0 to 10. 
This score reflects the overall quality of the response, based on five criteria:
1- Factual correctness.
2- Relevance to the question.
3- Proximity to the expected answer.
4- Hallucination avoidance.
5- Completeness.
The scoring prompt is provided in Figure~\ref{prompt_eval_score}.

\begin{table*}[ht!]
\centering
\resizebox{\textwidth}{!}{%
\begin{tabular}{lccccccccccccc}
\toprule
\multirow{3}{*}{\textbf{}} 
  & \multirow{3}{*}{\textbf{Video}}  
  & \multirow{3}{*}{\textbf{Subtitles}} 
  & \multirow{3}{*}{\makecell{\textbf{Meta}\\\textbf{data}}}
  & \multicolumn{4}{c}{\textbf{Grounding Skills}}
  & \multicolumn{4}{c}{\textbf{Reasoning Skills}}
  & \multirow{3}{*}{\makecell{\textbf{Avg.}\\\textbf{Acc.}}}
  & \multirow{3}{*}{\makecell{\textbf{Avg.}\\\textbf{Score}}} \\
\cmidrule(r){5-8} \cmidrule(r){9-12}
  &   &   &  
  & \makecell{\textbf{Global}\\\textbf{Appearance}}
  & \makecell{\textbf{Scene}\\\textbf{Transitions}}
  & \makecell{\textbf{Character}\\\textbf{Actions}}
  & \makecell{\textbf{Chronological}\\\textbf{Understanding}}
  & \makecell{\textbf{Summar-}\\\textbf{ization}}
  & \makecell{\textbf{Deep Context}\\\textbf{Understanding}}
  & \makecell{\textbf{Spoiler}\\\textbf{Understanding}}
  & \makecell{\textbf{Linking}\\\textbf{Events}}
  &   &   \\
\midrule
\rowcolor{gray!50}
Random Per. & -- & -- & -- & 20.0 & 20.0 & 20.0 & 20.0 & N/A & N/A & N/A & N/A & 20.0 & N/A \\

\multirow{3}{*}{w/o Filtering} & \cmark & \cmark & \xmark & 51.8 & 42.9 & 43.5 & 77.2 & 6.3 & 6.8 & 6.7 & 7.5 & 53.9 & 6.8 \\
\-  & \xmark & \cmark & \xmark & 29.1 & 39.6 & 46.8 & 82.0 & 7.2 & 7.0 & 5.9 & 7.5 & 49.4 & 6.9 \\
\-  & \xmark & \xmark & \cmark & 25.7 & 26.1 & 39.1 & 56.5 & 3.4 & 5.2 & 3.9 & 7.4 & 36.8 & 5.0 \\

\midrule
\multirow{3}{*}{w Filtering} & \cmark & \cmark & \xmark & 49.7 &	37.7 &	39.9 &	61.0 &	6.3 & 6.4 &	6.6 &	6.8 & 47.1 & 6.5 \\
\-  & \xmark & \cmark & \xmark & 26.5 & 34.2 & 42.2 & 68.5 & 7.1 & 6.5 & 5.8 & 6.8 & 42.8 & 6.5 \\
\-  & \xmark & \xmark & \cmark & 22.2 & 19.9 & 31.9 & 35.4 & 3.3 & 4.8 & 3.5 & 6.6 & 27.4 & 4.6 \\
\bottomrule
\end{tabular}
}
\caption{Analysis of the impact of filtering our benchmark by excluding the models' internal knowledge and emphasizing the visual questions. The reported numbers are on GPT4o. Table \ref{tab_appendix_blindness_experiment} in the Appendix, demonstrates the full results across more models. Random Per. is random performance.}
\label{tab:blindness_experiment}
\end{table*}

\begin{table*}[!ht]
    \centering
    \resizebox{\textwidth}{!}{%
    \begin{tabular}{lcccccccccccc}
    \toprule
\multirow{4}{*}{\textbf{Models}} & \multirow{4}{*}{\makecell{\textbf{Frame} \\ \textbf{Rate}}}  & \multicolumn{4}{c}{\textbf{Grounding Skills}} & \multicolumn{4}{c}{\textbf{Reasoning Skills}} & \multirow{4}{*}{\makecell{\textbf{Avg.} \\ \textbf{Acc.}}} & \multirow{4}{*}{\makecell{\textbf{Avg.} \\ \textbf{Score}}} & \multirow{4}{*}{\makecell{\textbf{} \textbf{Overall}}}\\ 
\cmidrule(r){3-6} \cmidrule(r){7-10}
\- & \-  & \multirow{2}{*}{\makecell{\textbf{Global} \\ \textbf{Appearance}}} & \multirow{2}{*}{\makecell{\textbf{Scene} \\ \textbf{Transitions}}} & \multirow{2}{*}{\makecell{\textbf{Character} \\ \textbf{Actions}}} & \multirow{2}{*}{\makecell{\textbf{Chronological} \\ \textbf{Understanding}}} & \multirow{2}{*}{\makecell{\textbf{Summar-} \\ \textbf{ization}}} & \multirow{2}{*}{\makecell{\textbf{Deep Context} \\ \textbf{Understanding}}} & \multirow{2}{*}{\makecell{\textbf{Spoiler} \\ \textbf{Understanding}}} & \multirow{2}{*}{\makecell{\textbf{Linking} \\ \textbf{Events}}}& \- & \-  \\
\- & \- & \- & \- & \- & \- & \- & \- & \- & \-  & \- & \-  & \-  \\

\cmidrule(r){1-1}
\cmidrule(r){2-2}
\cmidrule(r){3-6}
\cmidrule(r){7-10}
\cmidrule(r){11-12}
\cmidrule(r){13-13}

\rowcolor{gray!50} Random Per. & N/A & 20.0 & 20.0 & 20.0 & 20.0 & N/A & N/A & N/A & N/A & 20.0 & N/A &  N/A \\

GPT-4o & 450 FPV & \textbf{49.7} & 37.7 & 39.9 & \textbf{61.0} & \textbf{6.3} & \textbf{6.4} & \textbf{6.6} & \textbf{6.8} & \textbf{47.1} & \textbf{6.5} & \textbf{56.0} \\

Gemini-2.0 & 1 FPS & 45.1 & \textbf{39.3} & \textbf{50.0} & 50.1 & 5.7 & 6.0 & 4.3 & 5.4 & 46.1 & 5.4 & 49.9 \\

\midrule
InternVL3 & 128 FPV & \textbf{34.3} & 27.7 & 20.5 & 31.1 & \textbf{3.8} & 3.7 & 3.3 & 5.3 & 28.4 & 4.0 & \textbf{34.4} \\

Qwen2.5VL & 768 FPV & 30.0 & 25.3 & 22.7 & 20.3 & 3.3 & 4.3 & 3.4 & 5.4 & 24.6 & 4.1 & 32.8 \\

Qwen2VL & 768 FPV & 23.5 & 28.2 & 30.2 & 27.4 & 2.2 & 4.3 & \textbf{3.5} & 5.0 & 27.3 & 3.8 & 32.5 \\

Goldfish & 60 FPW & 16.2 & 22.9 & 21.3 & 25.4 & 3.0 & \textbf{4.9} & 3.4 & \textbf{5.6} & 21.5 & \textbf{4.2} & 31.9 \\

VideoChat-Flash & 1000 FPV & 20.5 & \textbf{29.5} & \textbf{34.9} & \textbf{37.4} & 2.6 & 3.4 & 2.2 & 4.2 & \textbf{30.6} & 3.1 & 30.9 \\

InternVL2 & 128 FPW & 26.6 & 24.9 & 21.5 & 26.6 & 2.9 & 3.5 & 3.0 & 5.0 & 24.9 & 3.6 & 30.4 \\

LLava-OneVision & 128 FPV & 21.0 & 24.7 & 23.3 & 30.3 & 2.0 & 3.7 & 2.7 & 5.1 & 24.8 & 3.4 & 29.4 \\

InternVL2.5 & 128 FPV & 27.1 & 25.1 & 21.2 & 29.2 & 2.4 & 2.8 & 2.1 & 4.2 & 25.6 & 2.9 & 27.4 \\

InternLM-XComposer & 16 FPW & 20.1 & 27.9 & 26.6 & 26.4 & 1.6 & 2.6 & 2.2 & 4.0 & 25.2 & 2.6 & 25.7 \\ 

LongVU & 512 FPV & 27.67 & 21.0 & 27.9 & 19.0 & 1.7 & 2.9 & 2.7 & 3.6 & 23.9 & 2.7 & 25.6 \\

MiniGPT4-video & 60 FPV & 18.1 & 23.1 & 25.9 & 23.1 & 2.0 & 2.9 & 2.0 & 3.3 & 22.5 & 2.6 & 24.1 \\

    \bottomrule
    \end{tabular}
    }
    \caption{{\papernameAbbrev} leaderboard across eight skills. Models are ranked by their equally weighted performance for Grounding and reasoning skills using this formula \(\left(0.5 \cdot \frac{\text{acc}}{100} + 0.5 \cdot \frac{\text{score}}{10}\right)\).
    FPV (Frames Per Video), FPS (Frames Per Second), and FPW (Frames Per Window) are reported. 
    All models use both video and subtitles as inputs. Random Per. is random performance.}
    \label{tab:final_results}
    \vspace{-3mm}
\end{table*}

\subsection{Findings and Insights}
We analyze the results across 13 models to uncover key patterns in long-video understanding. 
Our findings highlight substantial performance gaps, modality dependencies, and emerging limitations in current MLLMs, even at the frontier.

\subsubsection{World Knowledge \& Data Leakage}

To isolate the influence of pre-trained knowledge from visual understanding, we conducted a controlled experiment where models were prompted using only video metadata—such as movie titles, release years, or TV episode identifiers—without access to any visual or subtitle input.
As shown in Tables~\ref{tab:blindness_experiment} and~\ref{tab_appendix_blindness_experiment}, several models perform surprisingly well under this setting. 
For example, GPT-4o achieves 36.8\% and 5.0 in grounding and reasoning scores, respectively, indicating a substantial reliance on memorized knowledge from pretraining.
To mitigate this issue, we apply a filtering strategy described in Section~\ref{sec_method_data_integrity}. 
Post-filtering, model performance under metadata-only input drops significantly—e.g., GPT-4o’s accuracy decreases from 36.8\% to 27.4\%, approaching the random baseline of 20\%. 
This validates the effectiveness of our filtering and suggests that the remaining questions require genuine understanding of the visual and textual input.
Moreover, when full video and subtitle context is provided, model performance increases substantially (e.g., GPT-4o improves from 36.8\% to 53.9\%), reinforcing the role of multimodal input in successful reasoning.
This finding raises an important question:
\textit{\textbf{``Do current models inherently favor memorized knowledge over true multi-modal understanding?''}}
Our results suggest that overcoming this bias is key to advancing long-video reasoning.

\subsubsection{Models Are Struggling}

Despite recent progress in vision-language modeling, none of the evaluated models—commercial or open-source—achieve strong performance on grounding-based skills. 
As shown in Table~\ref{tab:final_results}, the best-performing model, GPT-4o, reaches only 47.1\% accuracy, approximately 2.5 times the random baseline of 20\%. 
This substantial gap highlights a key limitation: even state-of-the-art models struggle to maintain accurate visual grounding over long temporal contexts.
Reasoning-based performance is similarly low. 
GPT-4o achieves the highest reasoning score among commercial models with an average of only 6.5, followed by Gemini-2 at 5.4, and the best open-source model, InternVL3, at just 4.0.
These results underscore the difficulty of our benchmark and point to the need for further advancements in long video understanding and multi-step reasoning capabilities.

\subsubsection{Modality Benefits vs. Context Limitations}
To further analyze the contribution of each modality, we compare model performance across three input settings: 
(1) metadata only, 
(2) subtitles only, and 
(3) video combined with subtitles.
As shown in Tables~\ref{tab:blindness_experiment} and~\ref{tab_appendix_blindness_experiment}, subtitles alone offer only marginal improvements over metadata for certain visually grounded skills such as \textit{Global Appearance}, increasing accuracy from 22.2\% to 29.1\%. 
However, for multi-modal skills that require both visual and textual reasoning, e.g., \textit{Chronological Understanding}, the performance improves substantially, from 35.4\% to 68.5\%.
Surprisingly, adding video can be inconsistent: it significantly boosts performance on skills like \textit{Global Appearance} and \textit{Spoiler Questions}, but offers no benefit, or even degrades performance, for tasks such as \textit{Linking Events} and \textit{Character Actions}.
We hypothesize that these inconsistencies stem from the tension between visual diversity and context length. 
Many skills require long-range contextual reasoning, which subtitles provide in a continuous form. 
In contrast, video inputs, sampled uniformly across the full episode, introduce fragmented and scattered visual information that can overwhelm the limited context window of current models. 
This fragmentation may distract the model, leading to worse performance than when using subtitles alone.
This highlights a fundamental trade-off: while some skills benefit from multi-modal inputs, others rely more heavily on extended, coherent textual context, which can be disrupted by the inclusion of sampled visual frames.
Overall, the results underscore a key limitation in existing models: constrained context windows hinder their ability to jointly reason over extended textual narratives and dispersed visual evidence.

\subsubsection{Grounding \& Reasoning: Both Are Challenging}
Across the eight skill types, both grounding and reasoning tasks remain far from solved. 
While commercial models outperform open-source ones by a notable margin, all models struggle across both skill categories, with no single model dominating in either domain. 
This reflects the intrinsic difficulty of the benchmark and underscores the need for significant model innovation, particularly in handling extended temporal structure, entity tracking, entity linking, and causal reasoning. 
Our results suggest that long-video understanding is far from solved and demands focused future efforts.

\subsubsection{Impact of Different Frame Rates}
To assess the influence of input frame rates, we conducted an ablation study using four open-source models (QwenVL2.5, QwenVL2.0, InternVL3.0, and Video-Flash).
As shown in Table~\ref{tab:different_frame_rates}, increasing the number of input frames generally leads to better accuracy, although the degree of improvement varies between models.
For example, both QwenVL2.5 and QwenVL2.0 show substantial improvements with higher frame rates. On the MCQ set, QwenVL2.5 performance increases from 23.6\% to 27.4\%, while QwenVL2.0 improves from 24.6\% to 28.7\% as the number of frames increases from 16 to 128.
In contrast, Video-Flash and InternVL3.0 exhibit minimal improvement even when the number of frames is increased by a factor of eight in InternVL3.0 and by a factor of 62 in the case of VideoChat-Flash. 
The performance of QwenVL2.5 and QwenVL2.0 begins to degrade when entering the 768 frames, indicating that these models struggle when the context window is fully saturated.
A likely explanation is that, these models are constrained to a certain number of input frames, limiting their capacity to exploit longer temporal context.
In particular, reasoning-based skills benefit the most from higher frame rates, as the additional visual context reduces the likelihood of missing critical shots, thus improving model performance. 
In summary, while higher frame rates tend to improve accuracy, the degree of benefit is closely related to the architectural capacity of the model and the training strategy.

\begin{table*}[h!]
    \centering
    \resizebox{\textwidth}{!}{%
    \begin{tabular}{lccccccccccc}
    \toprule
\multirow{3}{*}{\textbf{Models}} & \multirow{3}{*}{\makecell{\textbf{Frame} \\ \textbf{Rate}}} & \multicolumn{4}{c}{\textbf{Grounding Skills}} & \multicolumn{4}{c}{\textbf{Reasoning Skills}} & \multirow{3}{*}{\makecell{\textbf{Avg.} \\ \textbf{Acc.}}} & \multirow{3}{*}{\makecell{\textbf{Avg.} \\ \textbf{Score}}}\\ 
\cmidrule(r){3-6} \cmidrule(r){7-10}
\- & \- & \multirow{2}{*}{\makecell{\textbf{Global} \\ \textbf{Appearance}}} & \multirow{2}{*}{\makecell{\textbf{Scene} \\ \textbf{Transitions}}} & \multirow{2}{*}{\makecell{\textbf{Character} \\ \textbf{Actions}}} & \multirow{2}{*}{\makecell{\textbf{Chronological} \\ \textbf{Understanding}}} & \multirow{2}{*}{\makecell{\textbf{Summar-} \\ \textbf{ization}}} & \multirow{2}{*}{\makecell{\textbf{Deep Context} \\ \textbf{Understanding}}} & \multirow{2}{*}{\makecell{\textbf{Spoiler} \\ \textbf{Understanding}}} & \multirow{2}{*}{\makecell{\textbf{Linking} \\ \textbf{Events}}}& \- & \-  \\
\- & \- & \- & \- & \- & \- & \- & \- \\

\cmidrule(r){1-1}
\cmidrule(r){2-2}
\cmidrule(r){3-6}
\cmidrule(r){7-10}
\cmidrule(r){11-12}

\rowcolor{gray!50} Baseline Random & N/A & 20.0 & 20.0 & 20.0 & 20.0 & N/A & N/A & N/A & N/A & 20.0 & N/A \\

\multirow{6}{*}{\textbf{QwenVL2.5}} & 768 & 30.0 & 25.3 & 22.7 & 20.3 & 3.3 & 4.3 & 3.4 & 5.4 & 24.6 & 4.1 \\
\-  & 400 & 31.3 & 27.1 & 24.0 & 23.3 & 3.5 & 4.3 & 3.3 & 5.4 & 26.4 & 4.1 \\
\-  &  350  & 30.9 & 27.8 & 24.1 & 23.9 & 3.4 & 4.2 & 3.7 & 5.6 & 26.6 & \textbf{4.2} \\
\-  &  256  & 33.3 & 29.8 & 24.6 & 23.8 & 3.3 & 4.1 & 3.4 & 5.4 & \textbf{27.9} & 4.1 \\
\-  &  128  & 32.0 & 29.7 & 22.1 & 25.8 & 2.9 & 3.7 & 3.3 & 5.2 & 27.4 & 3.7 \\
\-  &  16  & 28.3 & 25.1 & 21.2 & 19.8 & 2.2 & 3.2 & 2.9 & 4.8 & 23.6 & 3.3 \\
\midrule

\multirow{4}{*}{\textbf{QwenVL2.0}} & 768&23.5 & 28.2 & 30.2 & 27.4 & 2.2 & 4.3 & \textbf{3.5} & 5.0 & 27.3 & 3.8 \\
\-  &  256 & 27.5 & 30.9 & 30.2 & 29.9 & 2.2 & 4.0 & 3.1 & 5.1 & \textbf{29.6} & \textbf{3.6} \\
\-  &  128  & 27.9 & 29.0 & 29.2 & 28.9 & 2.2 & 3.7 & 3.4 & 4.8 & 28.7 & 3.5 \\
\-  &  16  & 26.4 & 22.6 & 23.9 & 25.4 & 1.8 & 3.0 & 3.1 & 4.6 & 24.6 & 3.1 \\
\midrule
\multirow{2}{*}{\textbf{InterVL3}} & 128 & 34.3 & 27.8 & 20.5 & 31.1 & 3.8 & 3.7 & 3.3 & 5.3 & \textbf{28.4} & \textbf{4.0} \\
\-  &  16  & 35.7 & 26.8 & 18.8 & 28.2 & 3.0 & 3.2 & 3.7 & 5.0 & 27.4 & 3.7 \\
\midrule
\multirow{3}{*}{\textbf{VideoChat-Flash}} & 1000 & 20.5 & \textbf{29.5} & \textbf{34.9} & \textbf{37.4} & 2.6 & 3.4 & 2.2 & 4.2 & \textbf{30.6} & 3.1 \\
\-  &  128 & 18.0 & 30.0 & 33.5 & 36.8 & 2.4 & 2.8 & 2.2 & 4.0 & \textbf{29.6} & \textbf{2.9}\\
\-  &  16  & 18.4 & 32.9 & 29.5 & 34.1 & 2.1 & 2.5 & 1.9 & 3.8 & 28.7 & 2.6 \\

\bottomrule
\end{tabular}
}
\caption{Impact of varying frame rates on model performance.}
\label{tab:different_frame_rates}
\vspace{-3mm}
\end{table*}


\subsection{Evaluation Robustness}

\subsubsection{Self-enhancement Bias}
The main results in Table~\ref{tab:final_results} use GPT-4o as the judge model, which raises the potential concern of self-enhancement bias~\cite{zheng2023judging}, since GPT-4o serves as both the task model and evaluator. However, we argue that such bias is unlikely in our setup due to the clear separation between generation and evaluation.
Specifically, during inference, the model accesses both video and subtitles to generate answers, whereas the judge model evaluates responses using only the generated text and reference answers. This separation in goals, modalities, and inputs minimizes the risk of any model-specific advantage transferring across tasks.

To further assess evaluation objectivity, we conducted a multi-judge study using two strong and diverse alternatives: Qwen2.5-VL-7B-Instruct (open-source) and Gemini 2.0-flash (commercial). Pairwise Pearson and Spearman correlations revealed strong agreement:
1) GPT-4o-mini \& Gemini 2.0-flash: Pearson 0.97, Spearman 0.94
2) GPT-4o-mini \& Qwen2.5-VL: Pearson 0.95, Spearman 0.92
We also report a Krippendorff’s alpha of 0.8014 across all judges, indicating high inter-rater reliability. These results demonstrate that GPT-4o is well-aligned with independent judges, supporting the robustness and impartiality of our evaluation.

\begin{table}[ht!]
\centering
\resizebox{\linewidth}{!}{
\begin{tabular}{lccc}
\toprule
\textbf{Model / Judge} & \textbf{GPT-4o} & \textbf{Gemini 2.0} & \textbf{Qwen 2.5} \\
\cmidrule(r){1-1} 
\cmidrule(r){2-4}
GPT-4o & 6.77 & 7.67 & 6.76 \\
Gemini-2.0 & 5.61 & 6.76 & 5.78 \\
Goldfish (Mistral) & 4.64 & 4.53 & 5.06 \\
Qwen2.5VL & 4.56 & 4.58 & 5.36 \\
InternVL3 & 4.46 & 4.89 & 5.21 \\
Qwen2VL & 4.11 & 4.00 & 4.50 \\
InternVL2 & 3.90 & 4.01 & 4.50 \\
LLava-Onevision & 3.89 & 3.84 & 4.38 \\
Video-Flash & 3.46 & 3.84 & 3.73 \\
InternVL2.5 & 3.26 & 3.29 & 3.97 \\
MiniGPT4-video (Mistral) & 3.22 & 2.65 & 3.16 \\
LongVU & 3.00 & 3.24 & 3.41 \\
InternLM-XComposer & 2.99 & 3.06 & 3.28 \\
\bottomrule
\end{tabular}
}
\caption{Ablation study demonstrates our LLM-based judge robustness.}
\label{tab_judge_robustness}
\end{table}

\subsubsection{Reliability of the reasoning skills scoring}

To assess the alignment between our LLM-based judge and human evaluators, we conducted a detailed comparison on 10\% of the test set using standard statistical measures.
Specifically, we collected ratings from three independent human annotators per sample and computed the average human score. 
We then measured the Pearson correlation between the LLM scores and the averaged human ratings, obtaining a score of 0.92, indicating a high degree of linear agreement.
Additionally, we report Spearman correlation of 0.89, confirming that the LLM’s ranking of outputs closely matches human preferences.

To contextualize this alignment, we computed inter-annotator agreement using Krippendorff’s alpha, which yielded a value of 0.86, showing that the LLM aligns with human consensus nearly as well as humans agree among themselves. We also measured per-annotator Pearson correlations with the LLM, which ranged from 0.88 to 0.93, further supporting the strong consistency between the LLM and individual human raters.
These results demonstrate that our LLM judge exhibits near-human-level alignment, validating its use as a reliable proxy for human evaluation in our setting.

\section{Conclusion}

Our evaluation of current Multimodal Large Language Models (MLLMs) on the {\papernameAbbrev} benchmark reveals key limitations in long-video understanding. 
Top models like GPT-4o perform modestly on grounding tasks and struggle with reasoning-based skills that require causal inference and context-aware understanding. Models show significant reliance on pre-trained knowledge, performing better with metadata alone, but improve substantially when provided full video and subtitle context, emphasizing the importance of multimodal inputs.
We also observe that visual input diversity often conflicts with the limited context windows of current models, affecting performance on certain tasks. Overall, {\papernameAbbrev} highlights critical gaps in existing models and directs future research toward improving visual grounding and multi-step reasoning for long-video comprehension.
\clearpage
\section{Limitations}
Despite its strengths, our question–answer generation pipeline currently depends on the availability of transcripts, limiting its use to movies and TV shows. To broaden applicability, we will develop a more general pipeline that can handle arbitrary videos without preexisting transcripts. We also plan to leverage our dense, skill‐based annotations to create multi‐video benchmarks, such as season‐level evaluations, that extend each skill across related clips. Furthermore, our training split can support retrieval‐based MLLM methods that focus on relevant frames (inspired by Video-XL~\citep{liu2025video}), and we aim to explore long-video reasoning approaches, such as those based on the GRPO algorithm~\cite{shao2024deepseekmathpushinglimitsmathematical}, using samples from our dataset.  

\section{Acknowledgment}
We would like to express our sincere gratitude to \textbf{Habib Slim} for his valuable assistance in preparing the visualizations accompanying this paper. His attention to detail and artistic insight significantly enhanced the overall quality and clarity of the work.

\bibliography{custom}
\clearpage

\clearpage
\appendix

\begin{table*}[ht!]
\centering
\resizebox{\textwidth}{!}{%
\begin{tabular}{llccccccccccccc}
\toprule
\multirow{3}{*}{\textbf{\-Models}}  & \multirow{3}{*}{\textbf{Filter}} 
  & \multirow{3}{*}{\textbf{Video}}  
  & \multirow{3}{*}{\textbf{Subtitles}} 
  & \multirow{3}{*}{\makecell{\textbf{Meta}\\\textbf{data}}}
  & \multicolumn{4}{c}{\textbf{Grounding Skills}}
  & \multicolumn{4}{c}{\textbf{Reasoning Skills}}
  & \multirow{3}{*}{\makecell{\textbf{Avg.}\\\textbf{Acc.}}}
  & \multirow{3}{*}{\makecell{\textbf{Avg.}\\\textbf{Score}}} \\
\cmidrule(r){6-9} \cmidrule(r){10-13}
\- &   &   &   &  
  & \makecell{\textbf{Global}\\\textbf{Appearance}}
  & \makecell{\textbf{Scene}\\\textbf{Transitions}}
  & \makecell{\textbf{Character}\\\textbf{Actions}}
  & \makecell{\textbf{Chronological}\\\textbf{Understanding}}
  & \makecell{\textbf{Summar-}\\\textbf{ization}}
  & \makecell{\textbf{Deep Context}\\\textbf{Understanding}}
  & \makecell{\textbf{Spoiler}\\\textbf{Understanding}}
  & \makecell{\textbf{Linking}\\\textbf{Events}}
  &   &   \\
\midrule
\rowcolor{gray!50}
\textbf{Random Per} & --& -- & -- & -- & 20.0 & 20.0 & 20.0 & 20.0 & N/A & N/A & N/A & N/A & 20.0 & N/A \\

\multirow{6}{*}{\textbf{GPT4o}} & \multirow{3}{*}{w/o Filtering} & \cmark & \cmark & \xmark & 51.8 & 42.9 & 43.5 & 77.2 & 6.3 & 6.8 & 6.7 & 7.5 & 53.9 & 6.8 \\
\- & \- & \xmark & \cmark & \xmark & 29.1 & 39.6 & 46.8 & 82.0 & 7.2 & 7.0 & 5.9 & 7.5 & 49.4 & 6.9 \\
\- & \- & \xmark & \xmark & \cmark & 25.7 & 26.1 & 39.1 & 56.5 & 3.4 & 5.2 & 3.9 & 7.4 & 36.8 & 5.0 \\
\cmidrule(r){2-15}
\- & \multirow{3}{*}{w Filtering} & \cmark & \cmark & \xmark & 49.7 &	37.7 &	39.9 &	61.0 &	6.3 & 6.4 &	6.6 &	6.8 & 47.1 & 6.5 \\
\- & \-  & \xmark & \cmark & \xmark & 26.5 & 34.2 & 42.2 & 68.5 & 7.1 & 6.5 & 5.8 & 6.8 & 42.8 & 6.5 \\
\- & \-  & \xmark & \xmark & \cmark & 22.2 & 19.9 & 31.9 & 35.4 & 3.3 & 4.8 & 3.5 & 6.6 & 27.4 & 4.6 \\
\midrule

\multirow{6}{*}{\textbf{Qwen2.5VL}} & \multirow{3}{*}{w/o Filtering} & \cmark & \cmark & \xmark & 32.8 & 30.1 & 30.0 & 41.4 & 3.4 & 4.8 & 3.7 & 6.5 & 33.6 & 4.6 \\
\- & \- & \xmark & \cmark & \xmark & 22.7 & 36.2 & 31.0 & 56.9 & 4.8 & 5.2 & 3.6 & 6.6 & 36.7 & 5.1 \\
\- & \- & \xmark & \xmark & \cmark & 19.0 & 26.3 & 28.9 & 43.5 & 2.2 & 3.7 & 3.5 & 6.2 & 29.4 & 3.9 \\
\cmidrule(r){2-15}
\- & \multirow{3}{*}{w Filtering} & \cmark & \cmark & \xmark & 30.0 & 25.3 & 22.7 & 20.4 & 3.3 & 4.3 & 3.4 & 5.4 & 24.6 & 4.1 \\
\- & \-  & \xmark & \cmark & \xmark & 18.4 & 24.2 & 21.5 & 23.9 & 4.2 & 4.1 & 3.7 & 5.4 & 21.9 & 4.3 \\
\- & \-  & \xmark & \xmark & \cmark & 15.7 & 19.6 & 19.8 & 17.2 & 2.2 & 3.4 & 3.1 & 4.9 & 18.1 & 3.4 \\

\midrule

\multirow{6}{*}{\textbf{InternVL 3.0}} & \multirow{3}{*}{w/o Filtering} & \cmark & \cmark & \xmark & 35.9 & 27.8 & 24.8 & 41.2 & 3.9 & 4.1 & 3.6 & 6.2 & 32.4 & 4.4 \\
\- & \- & \xmark & \cmark & \xmark & 30.8 & 34.3 & 24.8 & 46.5 & 4.2 & 4.2 & 3.8 & 6.4 & 34.0 & 4.7 \\
\- & \- & \xmark & \xmark & \cmark & 25.0 & 25.6 & 26.5 & 38.9 & 1.9 & 3.6 & 3.9 & 6.1 & 29.0 & 3.9 \\
\cmidrule(r){2-15}
\- & \multirow{3}{*}{w Filtering} & \cmark & \cmark & \xmark & 34.3 & 27.8 & 20.5 & 31.1 & 3.8 & 3.7 & 3.3 & 5.2 & 28.4 & 4.0 \\
\- & \-  & \xmark & \cmark & \xmark & 29.2 & 31.2 & 20.8 & 33.9 & 4.2 & 3.9 & 3.5 & 5.4 & 28.8 & 4.2 \\
\- & \-  & \xmark & \xmark & \cmark & 23.4 & 23.3 & 23.3 & 24.9 & 2.0 & 3.3 & 3.6 & 5.0 & 23.7 & 3.5 \\

\midrule

\multirow{6}{*}{\textbf{LlaVa-OneVision}} & \multirow{3}{*}{w/o Filtering} & \cmark & \cmark & \xmark & 23.2 & 27.6 & 27.0 & 41.1 & 2.0 & 4.0 & 3.2 & 6.1 & 29.7 & 3.9 \\
\- & \- & \xmark & \cmark & \xmark & 20.8 & 27.8 & 29.8 & 43.4 & 4.0 & 5.1 & 3.7 & 6.6 & 30.4 & 4.8 \\
\- & \- & \xmark & \xmark & \cmark & 16.7 & 27.8 & 28.5 & 39.3 & 0.8 & 3.7 & 3.4 & 5.8 & 28.1 & 3.4 \\
\cmidrule(r){2-15}
\- & \multirow{3}{*}{w Filtering} & \cmark & \cmark & \xmark & 21.1 & 24.7 & 23.3 & 30.3 & 2.0 & 3.7 & 2.7 & 5.1 & 24.8 & 3.4 \\
\- & \-  & \xmark & \cmark & \xmark & 18.5 & 24.5 & 25.7 & 28.9 & 3.9 & 4.5 & 3.3 & 5.6 & 24.4 & 4.3 \\
\- & \-  & \xmark & \xmark & \cmark & 14.8 & 24.5 & 26.2 & 28.4 & 0.8 & 3.4 & 2.9 & 4.8 & 23.5 & 3.0 \\
\bottomrule
\end{tabular}
}
\caption{Analysis of the impact of filtering our benchmark by excluding the models' internal knowledge and emphasizing the visual questions. Random Per. is random performance.}
\label{tab_appendix_blindness_experiment}
\end{table*}

\begin{table*}[!ht]
    \centering
    \resizebox{\textwidth}{!}{%
    \begin{tabular}{lcccccccccccc}
    \toprule
\multirow{4}{*}{\textbf{Models}} & \multirow{4}{*}{\makecell{\textbf{Frame} \\ \textbf{Rate}}}  & \multicolumn{4}{c}{\textbf{Grounding Skills}} & \multicolumn{4}{c}{\textbf{Reasoning Skills}} & \multirow{4}{*}{\makecell{\textbf{Avg.} \\ \textbf{Acc.}}} & \multirow{4}{*}{\makecell{\textbf{Avg.} \\ \textbf{Score}}} & \multirow{4}{*}{\makecell{\textbf{} \textbf{Overall}}}\\ 
\cmidrule(r){3-6} \cmidrule(r){7-10}
\- & \-  & \multirow{2}{*}{\makecell{\textbf{Global} \\ \textbf{Appearance}}} & \multirow{2}{*}{\makecell{\textbf{Scene} \\ \textbf{Transitions}}} & \multirow{2}{*}{\makecell{\textbf{Character} \\ \textbf{Actions}}} & \multirow{2}{*}{\makecell{\textbf{Chronological} \\ \textbf{Understanding}}} & \multirow{2}{*}{\makecell{\textbf{Summar-} \\ \textbf{ization}}} & \multirow{2}{*}{\makecell{\textbf{Deep Context} \\ \textbf{Understanding}}} & \multirow{2}{*}{\makecell{\textbf{Spoiler} \\ \textbf{Understanding}}} & \multirow{2}{*}{\makecell{\textbf{Linking} \\ \textbf{Events}}}& \- & \-  \\
\- & \- & \- & \- & \- & \- & \- & \- & \- & \-  & \- & \-  & \-  \\

\cmidrule(r){1-1}
\cmidrule(r){2-2}
\cmidrule(r){3-6}
\cmidrule(r){7-10}
\cmidrule(r){11-12}
\cmidrule(r){13-13}

\rowcolor{gray!50} Random Per. & N/A & 19.48 & 16.67 & 16.52 & 40.56 & N/A & N/A & N/A & N/A & 23.31 & N/A &  N/A \\
GPT-4o & 450 FPV &   54.26  &  50.0  &  43.96  &  49.66  &  6.2  &  6.49  &  5.61  &  7.00  &  49.47  &  6.33  &  56.36\\
Gemini-2.0 & 1 FPS &   50.39  &  50.79  &  56.61  &  36.60  &  5.70  &  5.40  &  3.41  &  5.94  &  48.60  &  5.11  &  49.86\\
\midrule
InternVL3 & 128 FPV &  33.33  &  28.57  &  23.46  &  36.29  &  3.91  &  3.79  &  3.41  &  5.28  &  30.41  &  4.10  &  35.69\\
Qwen2VL & 768 FPV &  31.01  &  23.81  &  32.59  &  48.41  &  2.26  &  4.23  &  2.93  &  5.14  &  33.96  &  3.64  &  35.18\\
Qwen2.5VL & 768 FPV &  37.98  &  22.22  &  22.53  &  27.17  &  3.22  &  4.13  &  3.67  &  5.34  &  27.48  &  4.09  &  34.19\\
LLava-OneVision & 128 FPV &  37.21  &  25.40  &  20.11  &  43.91  &  1.99  &  3.66  &  3.28  &  5.21  &  31.66  &  3.54  &  33.50\\
Goldfish & 60 FPW &   20.93  &  12.70  &  21.97  &  38.48  &  2.88  &  4.85  &  3.61  &  5.48  &  23.52  &  4.21  &  32.79 \\
InternVL2 & 128 FPV &  24.03  &  15.87  &  22.16  &  42.16  &  2.74  &  3.37  &  2.87  &  5.02  &  26.06  &  3.50  &  30.53\\
VideoChat-Flash & 1000 FPV &  28.68  &  20.63  &  33.33  &  29.73  &  2.45  &  3.24  &  2.35  &  4.20  &  28.09  &  3.06  &  29.35\\
InternVL2.5 & 128 FPV &   34.88  &  20.63  &  21.6  &  32.79  &  2.60  &  2.78  &  2.35  &  4.20  &  27.48  &  2.98  &  28.65\\
InternLM-XComposer & 16 FPW &  32.56  &  14.29  &  28.49  &  30.11  &  1.72  &  2.50  &  2.11  &  4.13  &  26.36  &  2.62  &  26.26\\ 
MiniGPT4-video & 60 FPV &  17.83  &  11.11  &  27.00  &  40.1  &  2.04  &  2.87  &  2.15  &  3.39  &  24.01  &  2.61  &  25.07\\
LongVU & 512 FPV & 20.93  &  15.87  &  21.42  &  29.86  &  1.70  &  2.84  &  3.07  &  3.53  &  22.02  &  2.79  &  24.94\\
    \bottomrule
    \end{tabular}
    }
    \caption{{\papernameAbbrev} leaderboard on the validation-set, across eight skills. Models are ranked by their equally weighted performance for Grounding and reasoning skills using this formula \(\left(0.5 \cdot \frac{\text{acc}}{100} + 0.5 \cdot \frac{\text{score}}{10}\right)\).
    FPV (Frames Per Video), FPS (Frames Per Second), and FPW (Frames Per Window) are reported. 
    All models use both video and subtitles as inputs. Random Per. is random performance.}
    \label{tab:validation_Set_results}
\end{table*}

\begin{table*}[!ht]
    \centering
    \resizebox{\textwidth}{!}{%
    \begin{tabular}{lcccccccccccc}
    \toprule
\multirow{4}{*}{\textbf{Models}} & \multirow{4}{*}{\makecell{\textbf{Frame} \\ \textbf{Rate}}}  & \multicolumn{4}{c}{\textbf{Grounding Skills}} & \multicolumn{4}{c}{\textbf{Reasoning Skills}} & \multirow{4}{*}{\makecell{\textbf{Avg.} \\ \textbf{Acc.}}} & \multirow{4}{*}{\makecell{\textbf{Avg.} \\ \textbf{Score}}} & \multirow{4}{*}{\makecell{\textbf{} \textbf{Overall}}}\\ 
\cmidrule(r){3-6} \cmidrule(r){7-10}
\- & \-  & \multirow{2}{*}{\makecell{\textbf{Global} \\ \textbf{Appearance}}} & \multirow{2}{*}{\makecell{\textbf{Scene} \\ \textbf{Transitions}}} & \multirow{2}{*}{\makecell{\textbf{Character} \\ \textbf{Actions}}} & \multirow{2}{*}{\makecell{\textbf{Chronological} \\ \textbf{Understanding}}} & \multirow{2}{*}{\makecell{\textbf{Summar-} \\ \textbf{ization}}} & \multirow{2}{*}{\makecell{\textbf{Deep Context} \\ \textbf{Understanding}}} & \multirow{2}{*}{\makecell{\textbf{Spoiler} \\ \textbf{Understanding}}} & \multirow{2}{*}{\makecell{\textbf{Linking} \\ \textbf{Events}}}& \- & \-  \\
\- & \- & \- & \- & \- & \- & \- & \- & \- & \-  & \- & \-  & \-  \\

\cmidrule(r){1-1}
\cmidrule(r){2-2}
\cmidrule(r){3-6}
\cmidrule(r){7-10}
\cmidrule(r){11-12}
\cmidrule(r){13-13}

\rowcolor{gray!50} Random Per. & N/A & 20.00 & 20.00 & 20.00 & 20.00 & N/A & N/A & N/A & N/A & 20.00 & N/A &  N/A \\
GPT-4o & 450 FPV & 51.83  &  42.91  &  43.49  &  77.15  &  6.32  &  6.82  &  6.73  &  7.47  &  53.85  &  6.84  &  61.10\\
Gemini-2.0 & 1 FPS &  46.73  &  44.59  &  55.94  &  67.62  &  5.80  &  6.56  &  4.36  &  6.19  &  53.72  &  5.73  &  55.50\\
\midrule
Qwen2.5VL & 768 FPV &  32.78  &  30.10  &  30.04  &  41.36  &  3.35  &  4.84  &  3.71  &  6.46  &  33.57  &  4.59  &  39.74\\
InternVL3 & 128 FPV &  35.93  &  27.76  &  24.75  &  41.17  &  3.86  &  4.06  &  3.57  &  6.19  &  32.40  &  4.42  &  38.30\\
Qwen2VL & 768 FPV &  25.37  &  31.37  &  34.91  &  36.68  &  2.23  &  4.87  &  3.59  &  6.03  &  32.08  &  4.18  &  36.94\\
Goldfish & 60 FPW &  17.28  &  24.45  &  24.03  &  29.32  &  3.02  &  5.45  &  3.61  &  6.56  &  23.77  &  4.66  &  35.19\\
VideoChat-Flash & 1000 FPV &  21.48  &  31.37  &  37.48  &  48.14  &  2.64  &  3.93  &  2.38  &  5.10  &  34.62  &  3.51  &  34.87\\
LLava-OneVision & 128 FPV &  23.15  &  27.57  &  27.04  &  41.07  &  2.04  &  4.04  &  3.18  &  6.12  &  29.71  &  3.85  &  34.08\\
InternVL2 & 128 FPV &  27.72  &  26.18  &  23.89  &  30.47  &  2.89  &  3.74  &  3.16  &  5.96  &  27.07  &  3.94  &  33.22\\
InternVL2.5 & 128 FPV &  28.58  &  27.10  &  25.18  &  34.00  &  2.46  &  3.10  &  2.30  &  5.10  &  28.72  &  3.24  &  30.56\\
InternLM-XComposer & 16 FPW &  22.53  &  30.33  &  30.33  &  34.48  &  1.63  &  2.86  &  2.43  &  5.02  &  29.42  &  2.99  &  29.63\\ 
LongVU & 512 FPV & 27.04  &  22.38  &  25.89  &  21.49  &  1.70  &  3.26  &  3.00  &  4.19  &  24.20  &  3.04  &  27.29\\
MiniGPT4-video & 60 FPV &  18.52  &  25.84  &  29.18  &  25.79  &  2.09  &  3.09  &  2.21  &  3.89  &  24.83  &  2.82  &  26.52 \\
    \bottomrule
    \end{tabular}
    }
    \caption{{\papernameAbbrev} before filtration, Section \ref{sec_method_data_integrity}, leaderboard across eight skills. Models are ranked by their equally weighted performance for Grounding and reasoning skills using this formula \(\left(0.5 \cdot \frac{\text{acc}}{100} + 0.5 \cdot \frac{\text{score}}{10}\right)\).
    FPV (Frames Per Video), FPS (Frames Per Second), and FPW (Frames Per Window) are reported. 
    All models use both video and subtitles as inputs. Random Per. is random performance.}
    \label{tab:before_filtering_results}
\end{table*}
\begin{table*}[h!]
\centering
\resizebox{0.8\textwidth}{!}{%
\begin{tabular}{lccccccccccccc}
\toprule
\multirow{2}{*}{\textbf{Models}} & \multirow{2}{*}{\textbf{\# Frames}} &  \multirow{2}{*}{\makecell{\textbf{I don't} \\ \textbf{know option}}} & \multicolumn{4}{c}{\textbf{Grounding Skills}} &\\ 
\cmidrule(r){4-7}&
\- & \- & 
\multirow{2}{*}{\makecell{\textbf{Global} \\ \textbf{Appearance}}} & \multirow{2}{*}{\makecell{\textbf{Scene} \\ \textbf{Transitions}}} & \multirow{2}{*}{\makecell{\textbf{Character} \\ \textbf{Actions}}} & \multirow{2}{*}{\makecell{\textbf{Chronological} \\ \textbf{Understanding}}}& \multirow{2}{*}{\makecell{\textbf{Avg.} \\ \textbf{Acc.}}} \\ 

\- & \- & \- & \- & \- & \- & \- \\

\cmidrule(r){1-1}
\cmidrule(r){2-2}
\cmidrule(r){3-3}
\cmidrule(r){4-7}
\cmidrule(r){8-8}

\multirow{2}{*}{\textbf{Qwen2.5-VL}}  &  \multirow{2}{*}{\textbf{768}}  &  \cmark  & 30.0 & 25.3 & 22.7 & 20.3 & 24.6 \\
\-  &  \-  &  \xmark  &  31.4 & 23.3 & 23.9 & 24.3 & \textbf{25.7} \\
\midrule

\multirow{2}{*}{\textbf{InternVL3.0}}  &  \multirow{2}{*}{\textbf{128}}  &  \cmark  & 34.3 & 27.8 & 20.5 & 31.1 & 28.4 \\
\-  &  \-  &  \xmark  & 35.5 & 30.9 & 22.9 & 33.5 & \textbf{30.7} \\
\midrule

\multirow{2}{*}{\textbf{Qwen2-VL}}  &  \multirow{2}{*}{\textbf{768}}   &  \cmark  &  23.5 & 28.2 & 30.2 & 27.4 & 27.3 \\
\-  &  \-   &  \xmark  & 25.2 & 31.8 & 29.9 & 32.3 & \textbf{29.8} \\
\midrule

\multirow{2}{*}{\textbf{Video-flash}}  &  \multirow{2}{*}{\textbf{1000}} &  \cmark  & 20.5 & 29.6 & 34.9 & 37.4 & \textbf{30.6} \\
\-  &  \-   &  \xmark  & 20.8 & 27.4 & 31.9 & 36.8 & 29.2 \\
\midrule

\multirow{2}{*}{\textbf{InternLM-XC}}  &  \multirow{2}{*}{\textbf{16 FPW}}  &  \cmark  & 20.1 & 27.9 & 26.6 & 26.4 & 25.3 \\
\-  &  \- &  \xmark  & 22.6 & 30.1 & 25.5 & 26.6 & \textbf{26.2} \\

\midrule
\multirow{2}{*}{\textbf{InternVL2.5}}  &  \multirow{2}{*}{\textbf{128}}  &  \cmark  & 27.1 & 25.1 & 21.2 & 29.2 & 25.6 \\
\-  &  \-  &  \xmark  & 27.7 & 27.7 & 22.2 & 28.9 & \textbf{26.7} \\
\midrule

\multirow{2}{*}{\textbf{LongVU}}  &  \multirow{2}{*}{\textbf{512}}  &  \cmark  & 27.7 & 21.0 & 28.0 & 19.0 & 23.9\\
\- &  \-  &  \xmark  & 27.7 & 23.5 & 29.5 & 23.5 & \textbf{26.0} \\

\bottomrule
\end{tabular}
}
\caption{Analysis of the impact of incorporating the ``I don't know'' option on model performance.}
\label{Tab:idontknow}
\end{table*}

\section{Extended Experimental Results}
\label{sec_appendix_Extended_Experiment}

\subsection{Results on the Validation-set}
We are organizing a challenge based on our benchmark; therefore, we will not release the test set.
Accordingly, the follow-up works that will leverage our benchmark can use the results reported on the validation set reported in Table \ref{tab:validation_Set_results}.
As shown in \ref{tab:validation_Set_results}, the same insights and conclusions are seen, similar to those reported on the test-set in Table \ref{tab:final_results}.

\subsection{Results on the Full Test-set Before Filtration}
As discussed in Section \ref{sec_method_data_integrity}, we have applied two types of filtration to make our benchmark more reliable and robust: 1) Filtering the internal model’s knowledge by removing the questions that can be answered from the metadata solely. 2) Making vision matters by removing the questions that can be answered from the subtitles only.
The main results reported in Tables \ref{tab:final_results} and \ref{tab:validation_Set_results} are reported on the filtered version of our benchmark.

To demonstrate the importance of the filtration process, we demonstrate the results on the test-set before filtration in Table \ref{tab:before_filtering_results}.
As shown in Table \ref{tab:before_filtering_results} and in comparison to Table \ref{tab:final_results}, all the models achieve a higher performance before filtration, due to the knowledge leakage, and the intensive dependency on the input text modality, subtitles. So by filtering these two factors, the performance of all the models drops significantly. 
We believe the filtered version is more reliable and truly assesses the model's capabilities.

\subsection{Impact of the ``I Don't Know'' Option}
As shown in Table~\ref{Tab:idontknow}, removing the ``I don't know'' option leads to an increase in model accuracy. Without this option, models are forced to select an answer, even when uncertain, which can occasionally result in correct guesses due to chance. 
In contrast, the inclusion of the ``I don't know'' option requires the model to explicitly acknowledge uncertainty when it lacks sufficient evidence to answer correctly, thereby increasing task difficulty and realism.
Additionally, excluding the ``I don't know'' option reduces the number of choices per question, which increases the expected performance of random guessing. This factor should be considered when comparing results across configurations with and without this option.

\section{{\papernameAbbrev} Further Details}
\subsection{More Clarification about the generated questions}
Leveraging the grounding information obtained during the verification process, we generated additional event-based questions. Specifically, we utilized the chronological events in conjunction with other grounding skills such as global appearance, scene transitions, and character actions to enrich the verified set. For example, in scene transitions skill, we can formulate systematic questions such as ``What are the scene transitions that occur before the event ``Ross watches TV with Rachel''?'' This is feasible because we can extract the relevant transitions occurring before the event timestamp from the complete set of scene transitions using temporal annotations. The same approach is applied with global appearance and character actions; we formulated questions that refer to moments immediately preceding or following specific events in the video. These questions are only added to the verified set.

\begin{figure}
    \centering
    \includegraphics[width=1\linewidth]{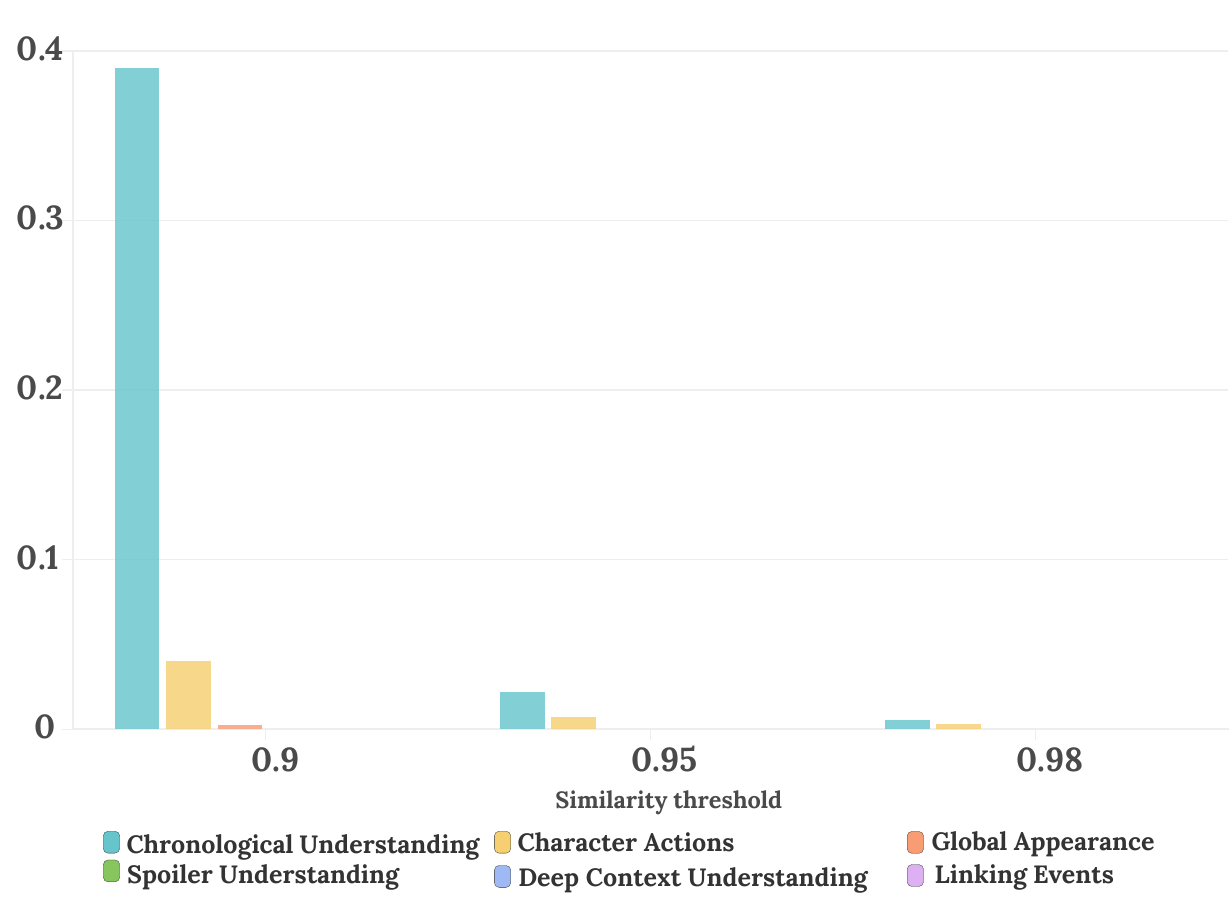}
    \caption{Duplication analysis for different thresholds using cosine similarity of text vector embeddings.}
    \label{fig:duplication_fig}
\end{figure}

\begin{figure*}[t!]
    \centering
    \includegraphics[width=0.9\textwidth]{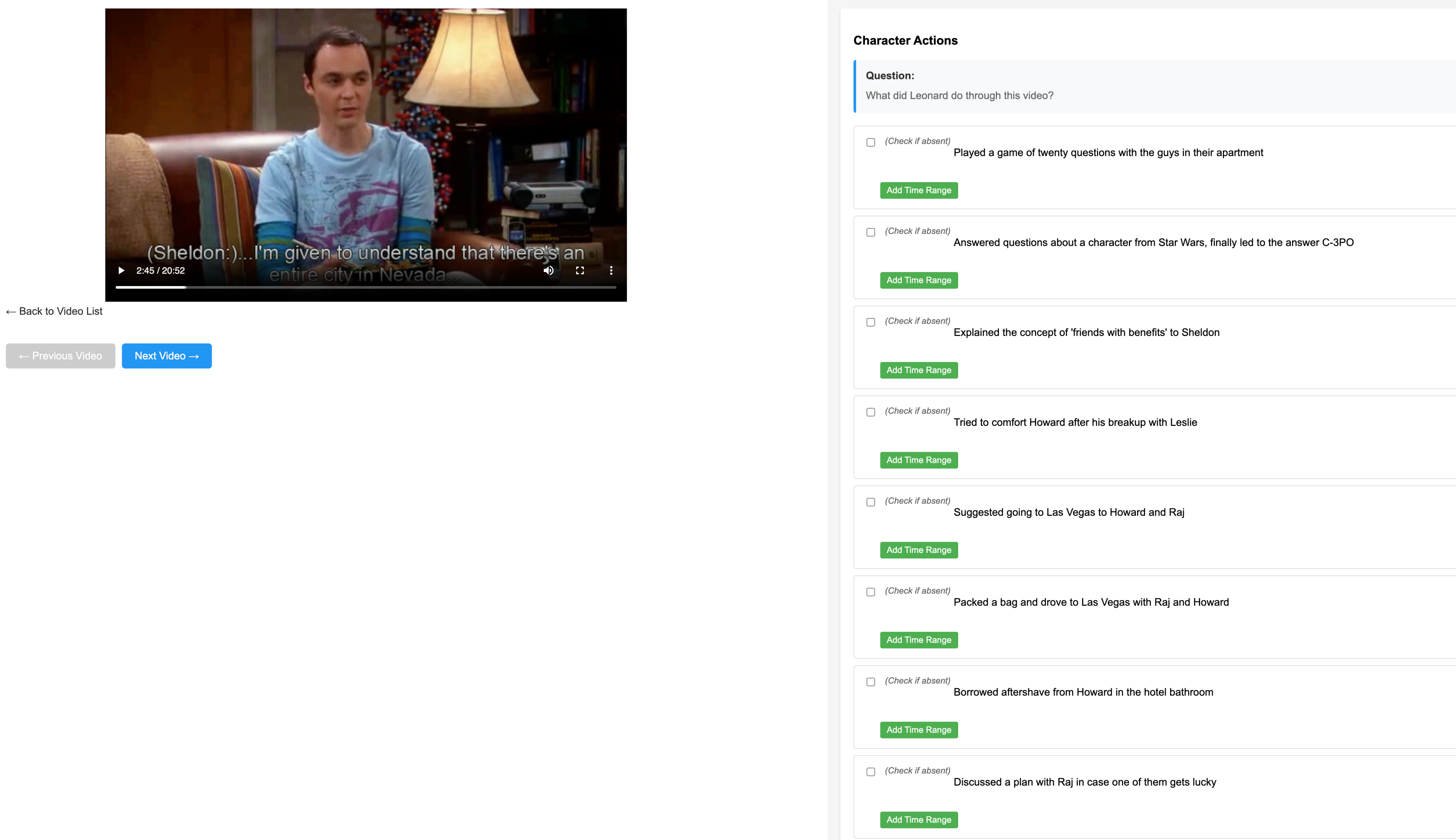}
    \caption{Data verification Website.}
    \label{fig:website_gui}
\end{figure*}

\subsection{Data Diversity}
\label{Data_Diversity}
Our benchmark is not narrowly scoped around human characters or interactions. Instead, it is built to holistically assess a wide range of long-video understanding capabilities:

\noindent \textbf{1. Diversity in Evaluated Skills}
Our benchmark explicitly targets a broad spectrum of cognitive and perceptual skills, not just character-centric ones. While certain tasks—such as Global Appearance and Character Actions—do emphasize characters, a substantial portion of the benchmark is designed around generic, high-level understanding of visual narratives, events, and environments. 
These include: Chronological Understanding, Deep-Context Understanding, Spoiler Questions, Linking Multiple Events, Scene Transitions.
These skills collectively test a model’s capability in grounding, temporal reasoning, and narrative comprehension far beyond surface-level character recognition.

\noindent \textbf{2. Diversity and Richness of Video Sources}
Our dataset comprises a large, balanced mix of TV shows and over 296 movies spanning multiple decades, cultures, genres, and visual settings. This includes diverse environments such as nature, abstract spaces, and sci-fi/fantasy worlds, demonstrating that the benchmark is not narrowly human-centric. 
For example :
Nature \& Survival: Movies like ``Into the Wild'' and ``Life of Pi'' focus on wilderness and animal interaction.
Abstract \& Visual Reasoning: Movies such as ``2001: A Space Odyssey'', ``Inception'', and ``Donnie Darko'' challenge models with surreal, symbolic visuals.
Space \& Environmental Focus: Movies such as "Gravity" and "The Thing" take place in isolated, non-social settings.
Sci-Fi \& Fantasy Worlds: Movies such as ``Avatar'' and ``The Matrix Reloaded'' emphasize non-human environments and abstract reasoning.
Minimal Dialogue / Object-Centric: ``Cube'', ``Moon'', and ``Minority Report'' require visual comprehension beyond character interaction.

Additionally, we argue that movies and TV shows are inherently more challenging than other long-form video domains, such as vlogs or surveillance footage. 
This is due to:
1) Multi-layered narratives require long-term memory and reasoning.
2) Non-linear structures involving flashbacks and fragmented timelines.
3) Visually rich environments that go beyond typical human-centric interactions.

These factors make them ideal testbeds for evaluating true long-video understanding.

\subsection{Data Copyrights}
\label{copyrights}
Our benchmark provides only annotations and uses two widely adopted, publicly available video datasets that provide legally compliant, downsampled video content (e.g., 3 FPS, scene shots). Specifically, we use: (1) six full-season TV shows from the TVQA~\cite{lei2019tvqa} dataset, based on the long-form version in GoldFish~\cite{ataallah2024goldfishvisionlanguageunderstandingarbitrarily}, and (2) movies from the MovieNet~\cite{huang2020movienet} dataset. For the TV shows, transcripts were sourced from reputable platforms such as foreverdreaming and manually verified by the authors for alignment with the video. This usage is consistent with fair use for academic research. For movies, we use transcripts provided by MovieNet~\cite{huang2020movienet}, which are already filtered and accurately aligned with dialogue and subtitles.

\subsection{Transcripts vs. Subtitles}
\label{sec:diff_trans_sub}
\begin{figure}
    \centering
    \includegraphics[width=1\linewidth]{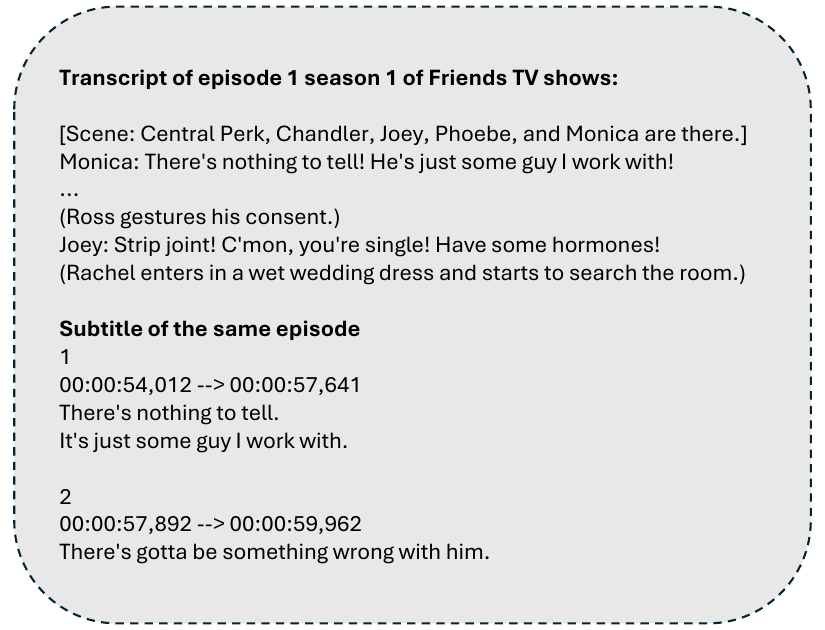}
    \caption{Difference between transcript and subtitles.}
    \label{fig:diff_trans_sub}
\end{figure}

Figure~\ref{fig:diff_trans_sub} illustrates the distinction between transcripts and subtitles. 

\textbf{Transcripts} are comprehensive documents authored by screenwriters, containing not only spoken dialogue but also detailed scene descriptions, setting and location information, character actions, and camera directions (e.g., angles, cuts, and shot composition). Effectively, transcripts serve as blueprints for both the visual and narrative structure of a movie or TV show. In our benchmark, transcripts are used exclusively during the data construction phase to generate rich and diverse questions. They are \emph{not} provided to the model during inference or evaluation.

\textbf{Subtitles}, by contrast, are limited to the spoken dialogue extracted from the video’s audio track. These are optionally provided to the model at inference time, serving as an auxiliary textual modality alongside the video frames. If available, subtitles can enrich the input context and help improve performance, particularly for dialogue-centric questions.

\subsection{Duplications Filtering}
\label{sec_appendix_duplication_filtering}
To ensure that our dataset is free of duplicate or near-duplicate questions, we implemented a semantic similarity filtering pipeline.

We first embedded each question and its answer choices using M3-Embedding~\citep{bge-m3}, then computed pairwise cosine similarity to detect potential duplicates. We experimented with similarity thresholds of 90\%, 95\%, and 98\%.
As shown in Figure~\ref{fig:duplication_fig}, the vast majority of questions are unique across all skill categories. Only two skills—\textit{Temporal Questions} and \textit{Character Actions}—produced potential duplicates. However, upon manual inspection, these were found to be false positives. 
For example, the following two questions were flagged as duplicates due to high embedding similarity, despite a critical semantic difference:

\begin{tcolorbox}[colback=gray!5!white, colframe=gray!80!black, boxrule=0.3pt, left=1pt, right=1pt, top=2pt, bottom=2pt]
\small
\textbf{Q1:} Did the event flashback to Phoebe completing a mile on a hippity-hop before turning thirty happen \textbf{before} the event Monica makes breakfast with chocolate-chip pancakes?\\
\textbf{Q2:} Did the event flashback to Phoebe completing a mile on a hippity-hop before turning thirty happen \textbf{after} the event Monica makes breakfast with chocolate-chip pancakes?
\end{tcolorbox}

Although these questions differ by a single word, that word fully reverses the temporal meaning, highlighting the importance of semantic context when evaluating similarity.
Based on this analysis, we conclude that the dataset contains no true duplicates. The filtering pipeline successfully identifies and flags near-duplicates, and the observed false positives illustrate the subtlety and complexity of semantic variation in question phrasing.

\subsection{Human Verification}
\label{sec_appendix_human_verf}

To ensure the reliability of our benchmark, we conducted a human evaluation on 20\% of the dataset. 
Annotators were tasked with verifying the correctness of the question–answer (Q/A) pairs across each skill category.
Figure~\ref{fig:website_gui} shows the custom-built verification interface developed to streamline the annotation process.

\noindent During verification, annotators assessed each video using the following criteria:

\begin{itemize}
  \item \textbf{Global Appearance and Character Actions:}  
    Annotators were provided with the character’s name and a list of appearance or action events extracted by our pipeline. Each event was marked as \emph{absent} if not observed in the video; if present, the annotator recorded its corresponding timestamp.

  \item \textbf{Linking Events and Deep Context Understanding:}  
    Annotators confirmed the validity of each Q/A pair, ensuring the answer could only be derived from the visual or textual content of the video. Temporal grounding was also recorded when applicable.

  \item \textbf{Scene Transitions and Chronological Understanding:}  
    Annotators reviewed the sequence of collected events for each video, verifying their presence and providing timestamps for valid instances.

  \item \textbf{Spoiler Questions:}  
    Annotators ensured that each Q/A pair relied solely on information present in the video, excluding any external or prior knowledge.
\end{itemize}

\noindent The collected timestamps were cross-referenced with the dataset to validate both the extracted events and their chronological consistency. This also enabled the generation of \textit{temporal certificates} for each Q/A pair, following the methodology introduced in EgoSchema \cite{mangalam2023egoschema}. A temporal certificate represents the number of video segments required to answer a given question. Following \cite{mangalam2023egoschema}, video clips shorter than 0.1 seconds were discarded, and consecutive segments within a 5-second window were merged into a single clip. The total duration of these merged segments defines the certificate length.

Our human-verified test set contains an average of 3.23 temporal certificates per Q/A pair, with an average certificate length of 202.97 seconds, underscoring the benchmark’s emphasis on comprehensive long-video understanding. The presence of timestamps for each answer further enhances the traceability and trustworthiness of the verified set.

Accuracy statistics across skill categories are reported in Table~\ref{tab:human_verficaton}, with an overall verification accuracy of 90.11\%. Incorrect or ambiguous Q/A pairs were removed, and only validated examples were retained for the final test set.

Note that \textit{Summarization} questions do not require human verification, as they are sourced from the official IMDb website \cite{imdb}. According to IMDb’s contribution \href{https://help.imdb.com/article/contribution/titles/plots/G56STCKTK7ESG7CP#}{guidelines}, all plot summaries must adhere to strict formatting and factual standards, and each submission is manually reviewed and approved by the IMDb team. This ensures the summaries are reliable, accurate, and effectively pre-verified by human annotators.

\begin{table}[t!]
\centering
\resizebox{0.48\textwidth}{!}{%
\begin{tabular}{lcc}
\toprule
\textbf{Skill Name} & \textbf{Number of Questions} & \textbf{Per Question Acc.(\%)} \\
\midrule
Character Actions & 1370 & 89.76 \\
Global Appearance & 750 & 93.21 \\ 
Scene Transitions & 224 & 85.23 \\ 
Chronological understandings & 224 & 90.37 \\
Deep Context Understanding & 5574 & 94.14 \\ 
Linking Events & 4587 & 95.92 \\ 
Spoiler Questions & 151 & 82.11 \\ 
\bottomrule
\end{tabular}
}
\caption{Human verification for the keypoints used in the generation pipeline. The number of questions corresponds to the ones shown to the annotators, not the final count in our dataset.}
\label{tab:human_verficaton}
\end{table}



\subsection{Models Configuration}
\label{sec:models_config}

\noindent\textbf{GPT-4o~\cite{GPT-4o}:}  
GPT-4o does not support direct ingestion of \texttt{.mp4} video files. To accommodate this, we sample up to 450 frames per video and provide the model with the corresponding subtitle text and question.

\noindent\textbf{Gemini-Flash 2.0~\cite{gemini_flash}:}  
Developed by Google, Gemini-Flash 2.0 supports native video input. For TV shows, we supply the model with the full video and the associated question. For movies—which lack audio—we additionally provide the subtitle file alongside the video and question.

\noindent\textbf{Qwen2.5-VL~\cite{bai2025qwen25vltechnicalreport} and Qwen2-VL~\cite{wang2024qwen2vlenhancingvisionlanguagemodels}:}  
These models accept up to 768 input frames, fitting within the memory limits of an NVIDIA A100 GPU (80~GB) under the authors' recommended inference settings. Subtitles are concatenated with the input question to provide multimodal context.

\noindent\textbf{Video-Chat-Flash~\cite{li2025videochatflashhierarchicalcompressionlongcontext}:}  
Although the model supports up to 10,000 frames, we limit input to 1,000 frames due to memory constraints on the A100 GPU (80~GB). Other configurations follow the default setup. Subtitles are concatenated with the input question to enrich contextual understanding.

\noindent\textbf{InternVL3.0~\cite{zhu2025internvl3exploringadvancedtraining}, InternVL2.5~\cite{InternVL2.5}, and InternVL2~\cite{InternVL2}:}  
For the InternVL family, we evaluate models using both 16 and 128 frames, the latter being the maximum supported without hallucination on an A100 GPU. Although these models can technically process more frames, exceeding 128 often introduces unstable outputs. Subtitles corresponding to the selected frames are appended to the input question.

\noindent\textbf{LLaVA-OneVision~\cite{li2024llavaonevisioneasyvisualtask}:}  
This model supports both image and video inputs. We evaluate performance at 16 and 128 frames—found to be the upper bound for stable performance on an A100 GPU. Subtitles for selected frames are appended to the question.

\noindent\textbf{LongVU~\cite{shen2024longvuspatiotemporaladaptivecompression}:}  
LongVU is evaluated using 512 frames, staying within the memory and context window limits of the A100 GPU. Subtitles aligned with the selected frames are appended to the input question.

\noindent\textbf{InternLM-XComposer-2.5~\cite{zhang2024internlmxcomposer25versatilelargevision}:}  
We use the default configuration with a 16-frame window and increase the number of clips per video to 120. Subtitles aligned with these clips are appended to the input question. All evaluations are conducted on an A100 GPU.

\noindent\textbf{Goldfish and MiniGPT-4-Video~\cite{ataallah2024goldfishvisionlanguageunderstandingarbitrarily}:}  
We use the Mistral backbone with default parameters, retrieving $k=3$ video clips per query. Each 80-second video yields 60 frames (approx. 1.3 FPS). Although the model supports up to 90 frames, we limit it to 60 to reserve space for the question, subtitles, and output tokens—reducing hallucinations due to context overflow. Subtitles are interleaved with the frames, following the MiniGPT-4-Video configuration.
For MiniGPT-4-Video, we similarly sample 60 frames per video, interleave them with subtitles, and prompt the model with the question for generation.



\subsection{Benchmark Examples}
\label{Benchmark_skills_examples}
\begin{figure*}[ht]
    \centering
    \includegraphics[width=1\textwidth]{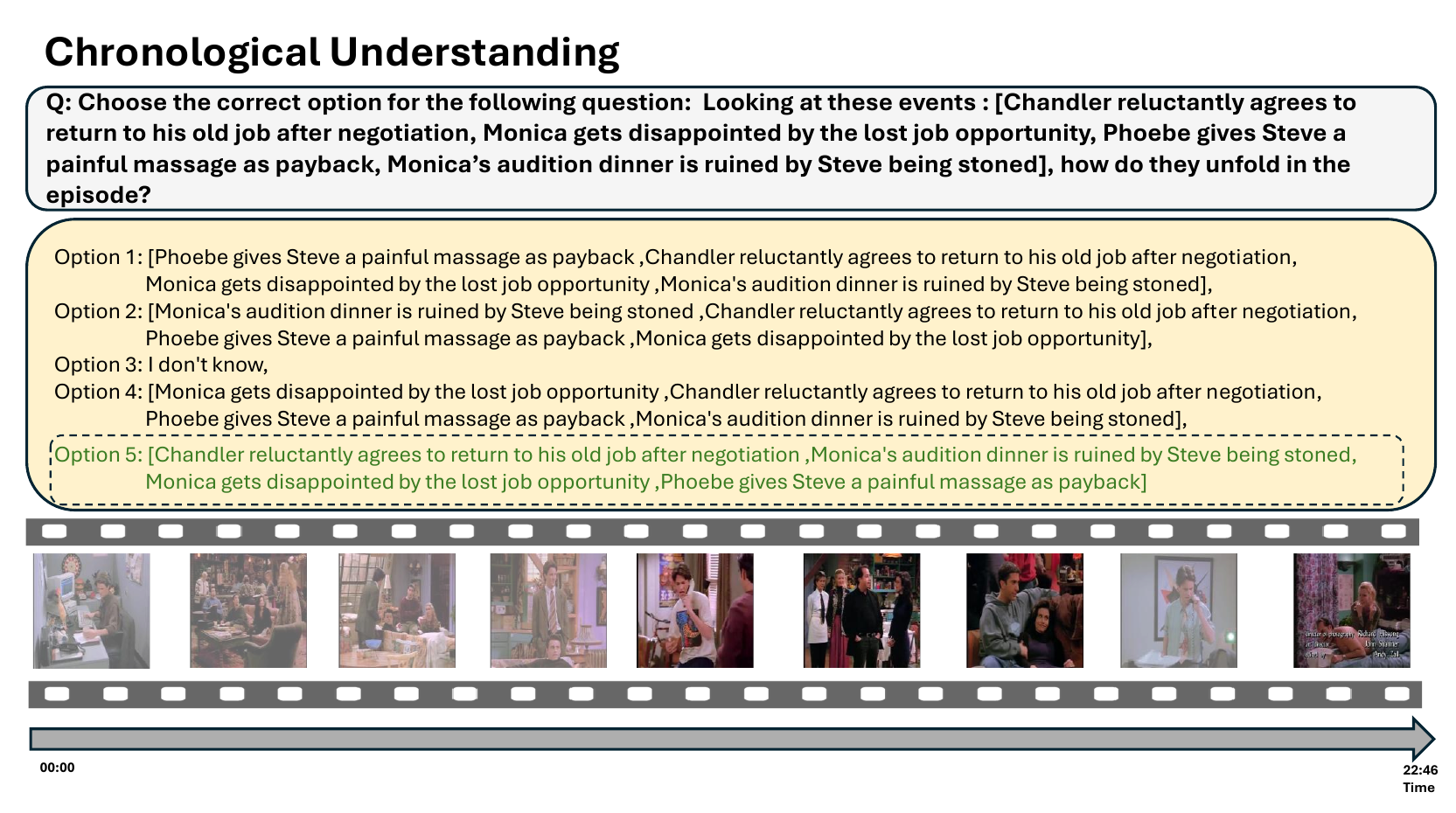}
    \caption{Example for the chronological understandings skill.}
    \label{temporal_questions}
\end{figure*}

\begin{figure*}[ht]
    \centering
    \includegraphics[width=1\textwidth]{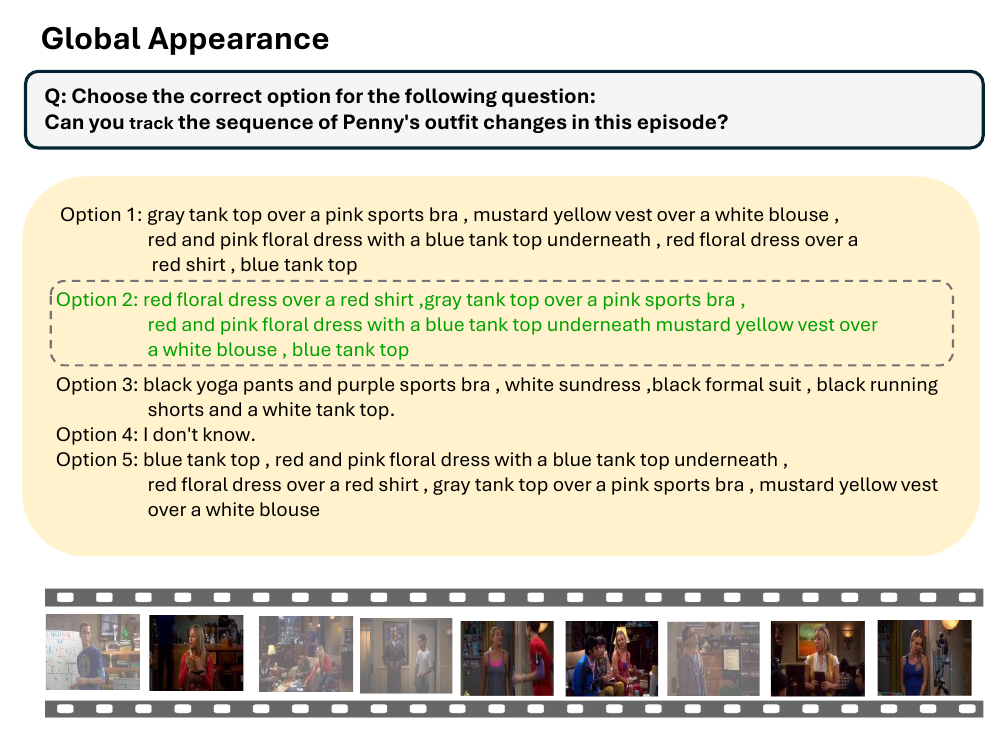}
    \caption{Example for the Global appearance skill.}
    \label{global_apperance}
\end{figure*}

\begin{figure*}[ht]
    \centering
    \includegraphics[width=1\textwidth]{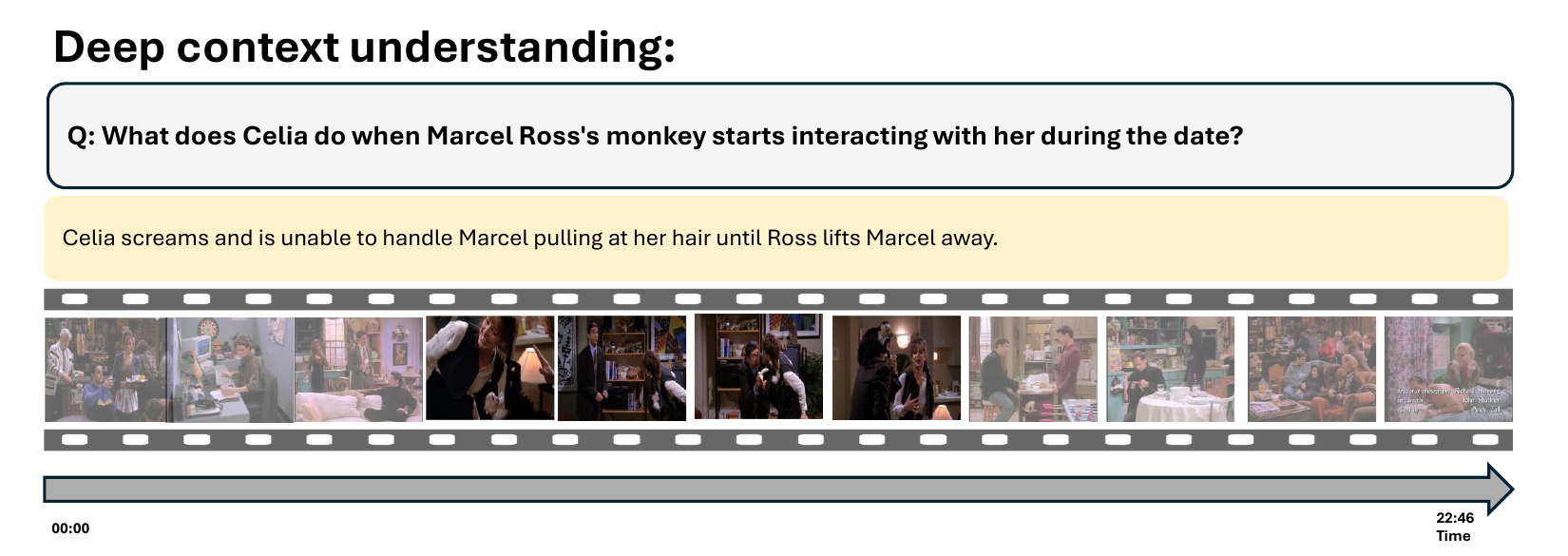}
    \caption{Example for the deep context understanding skill.}
    \label{context_understanding}
\end{figure*}

\begin{figure*}[ht]
    \centering
    \includegraphics[width=1\textwidth]{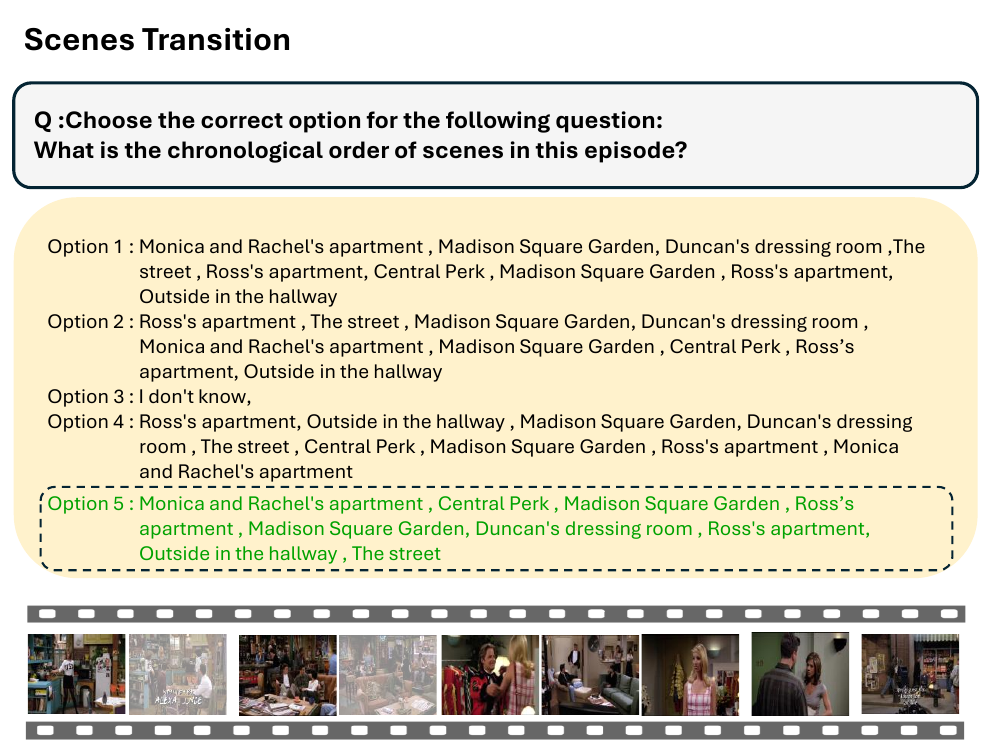}
    \caption{Example for the scenes transitions skills.}
    \label{scene_transition}
\end{figure*}

\begin{figure*}[ht]
    \centering
    \includegraphics[width=1\textwidth]{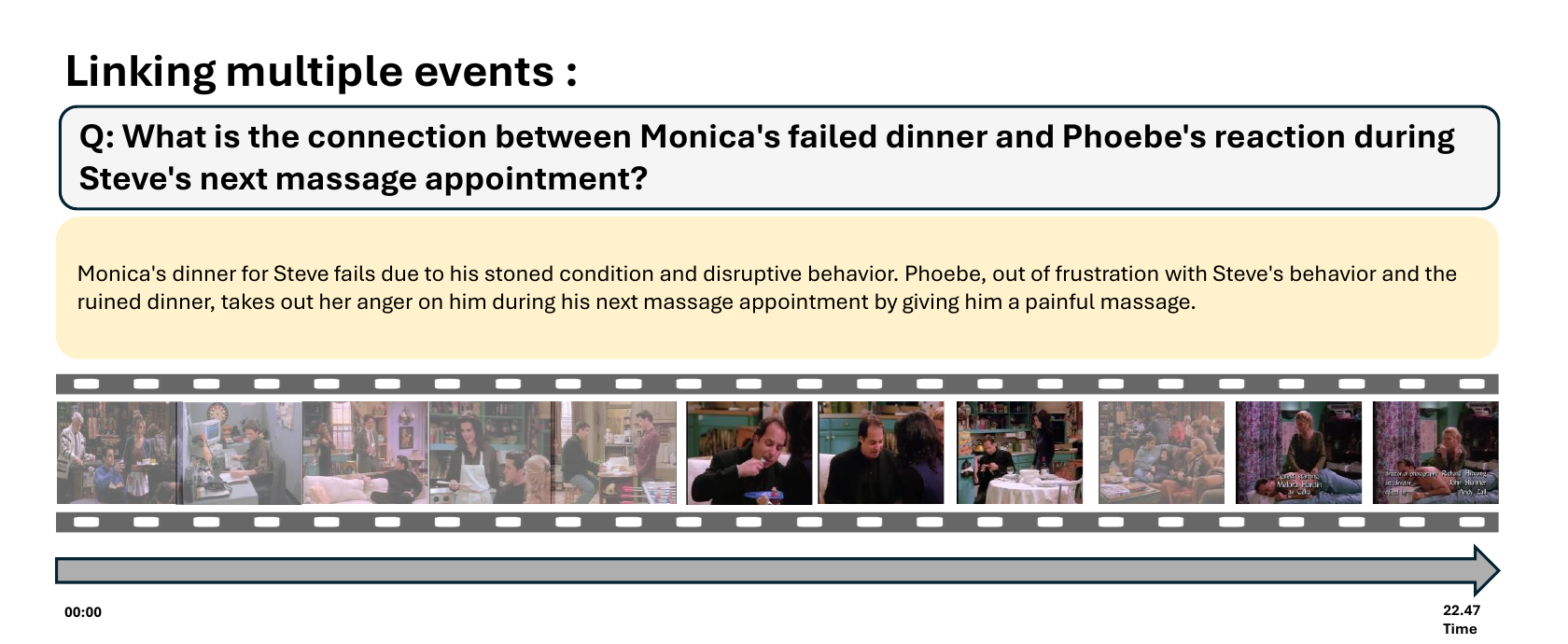}
    \caption{Example for the linking multiple events skill.}
    \label{linking_events}
\end{figure*}

\begin{figure*}[ht]
    \centering
    \includegraphics[width=1\textwidth]{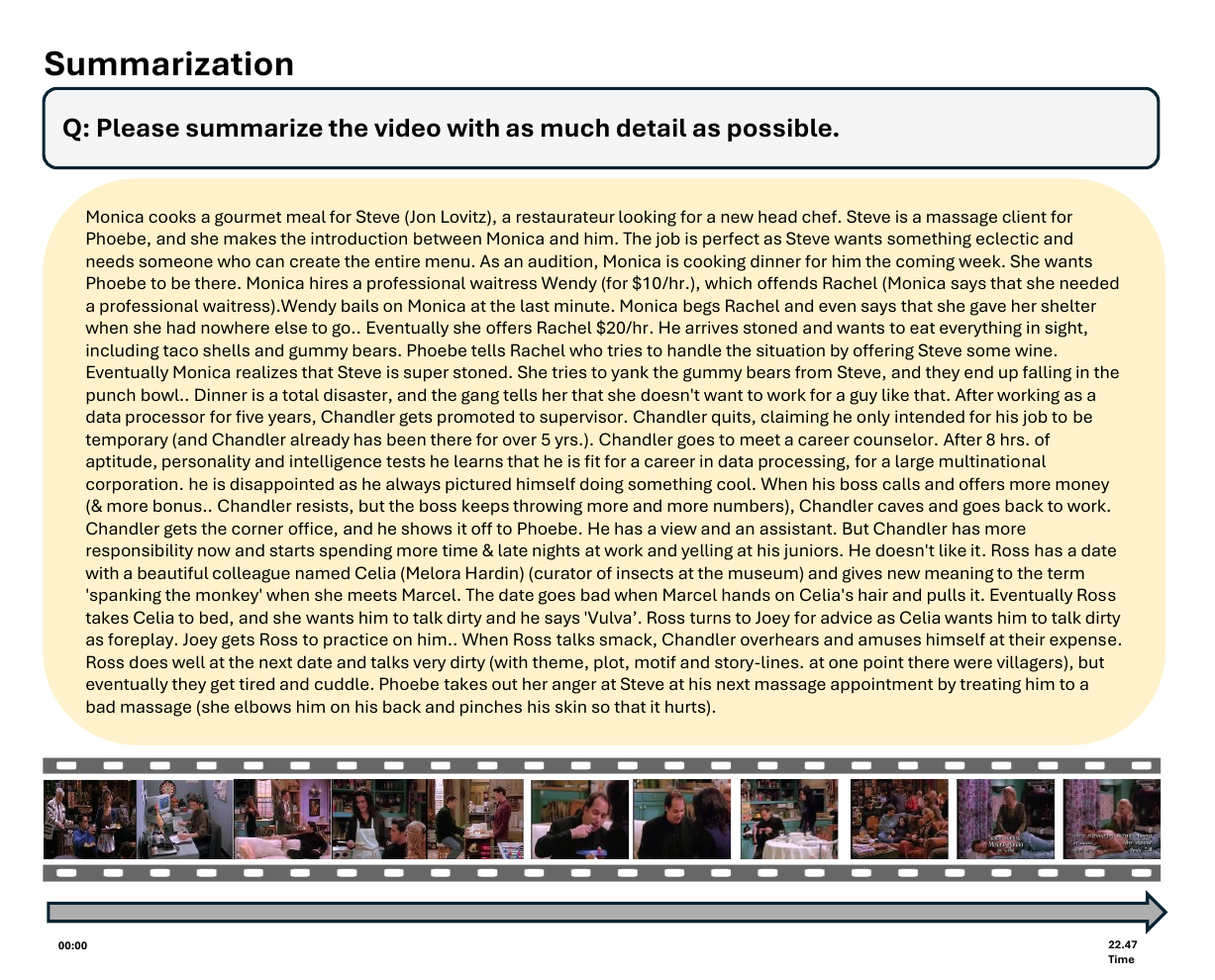}
    \caption{Example for the summarization skill.}
    \label{summarization}
\end{figure*}

\begin{figure*}[ht]
    \centering
    \includegraphics[width=1\textwidth]{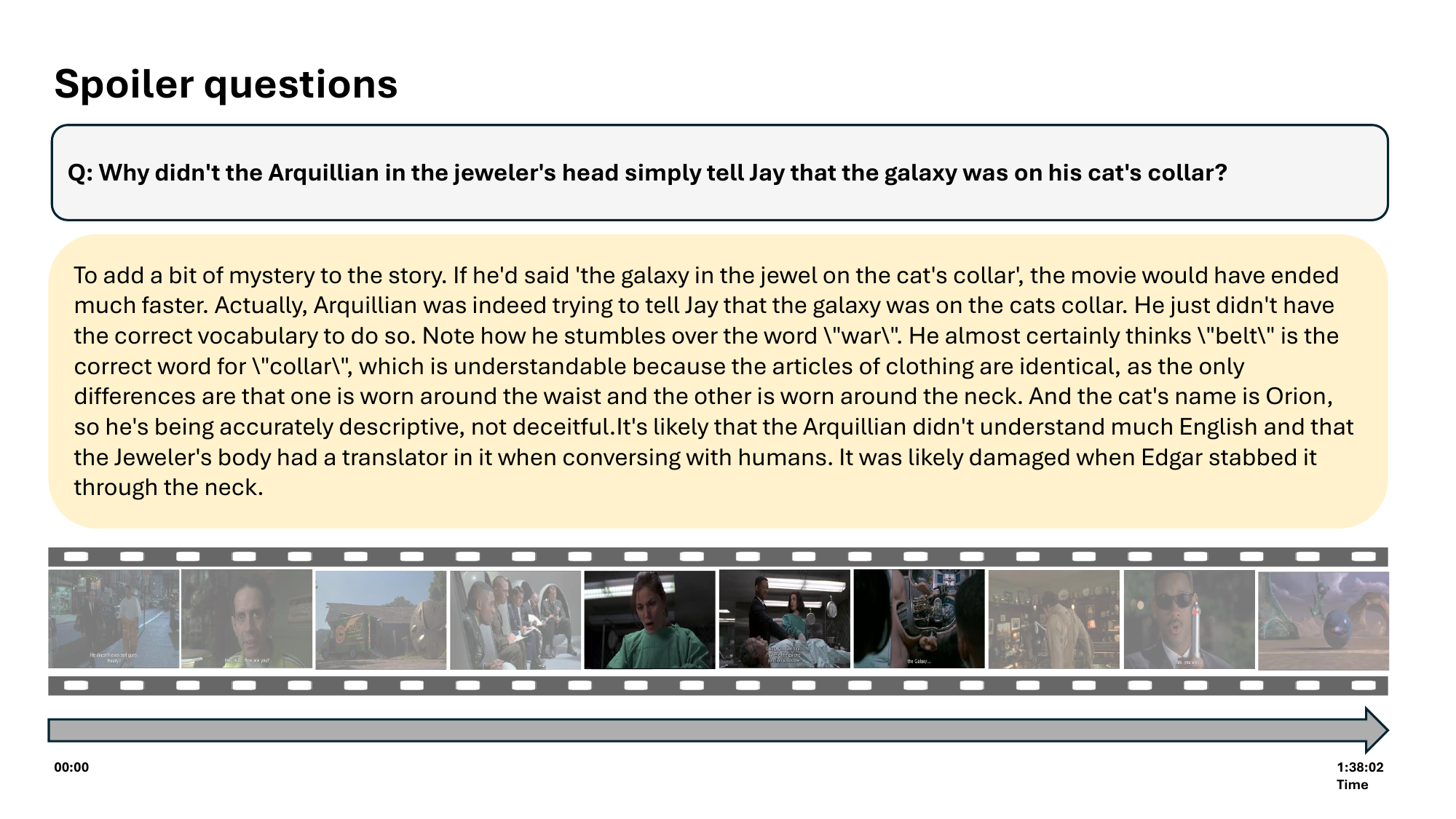}
    \caption{Example for the spoiler questions in Movies.}
    \label{spoiler_questions_example}
\end{figure*}

\begin{figure*}[ht]
    \centering
    \includegraphics[width=1\textwidth]{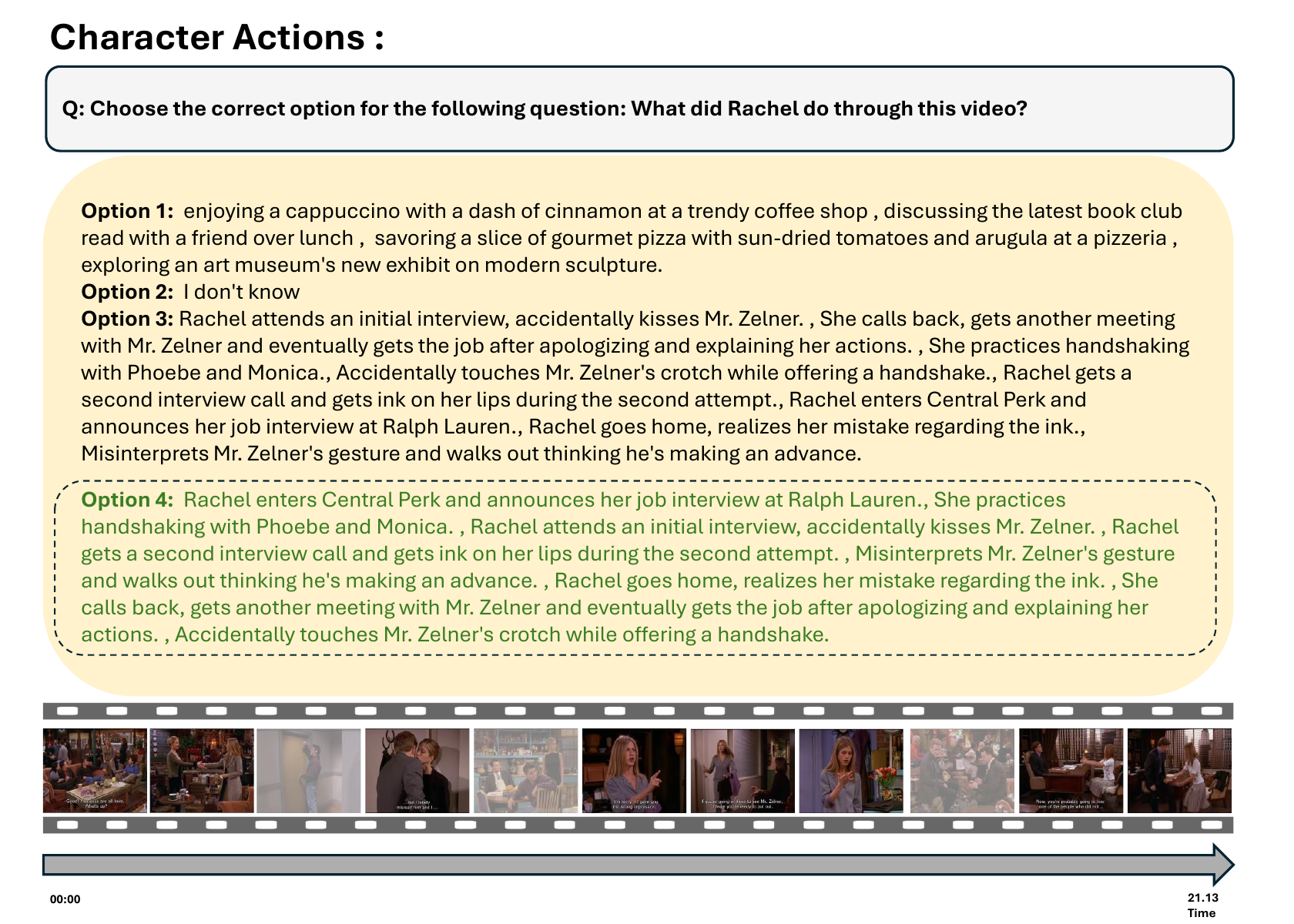}
    \caption{Example for character actions questions.}
    \label{character_actions_example}
\end{figure*}

Here in this section, we are showing more examples of our benchmark skills, such as the Chronological Understanding in Fig. \ref{temporal_questions}, linking multiple events in Figure.\ref{linking_events}, and deep context understanding in Figure. \ref{context_understanding}, spoiler questions in Figure.\ref{spoiler_questions_example}, Character Actions Figure.\ref{character_actions_example},Scene transitions in Figure. \ref{scene_transition}, Global Appearance in Figure \ref{global_apperance}, and summarization in Figure. \ref{summarization}.

\subsection{Success and Failure Cases}
\begin{figure*}[ht]
    \centering
    \includegraphics[width=1\textwidth]{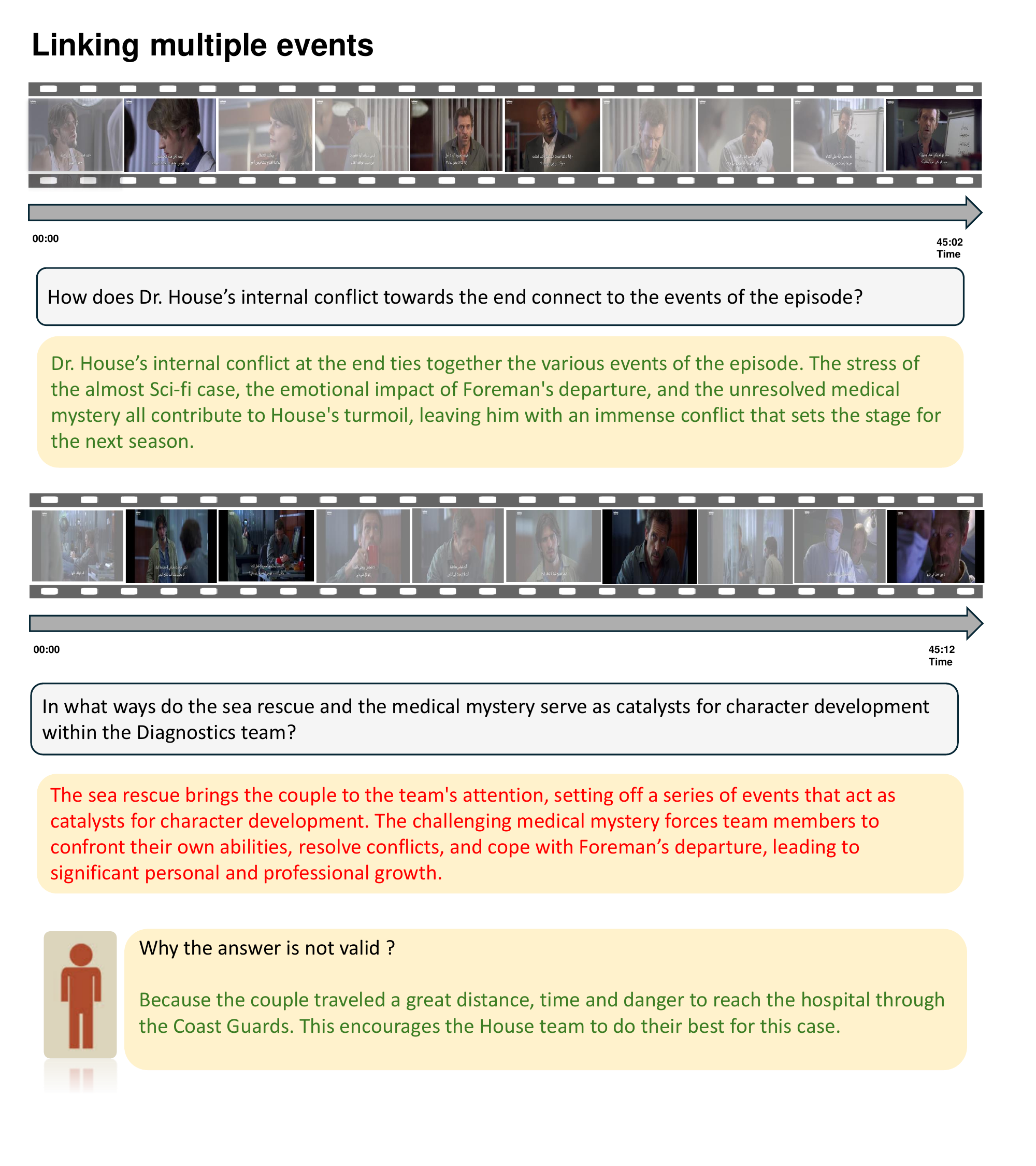}
    \caption{Examples of success and failure cases in Linking Multiple Events questions.}
    \label{linking_failure_and_success}
\end{figure*}

\begin{figure*}[ht]
    \centering
    \includegraphics[width=1\textwidth]{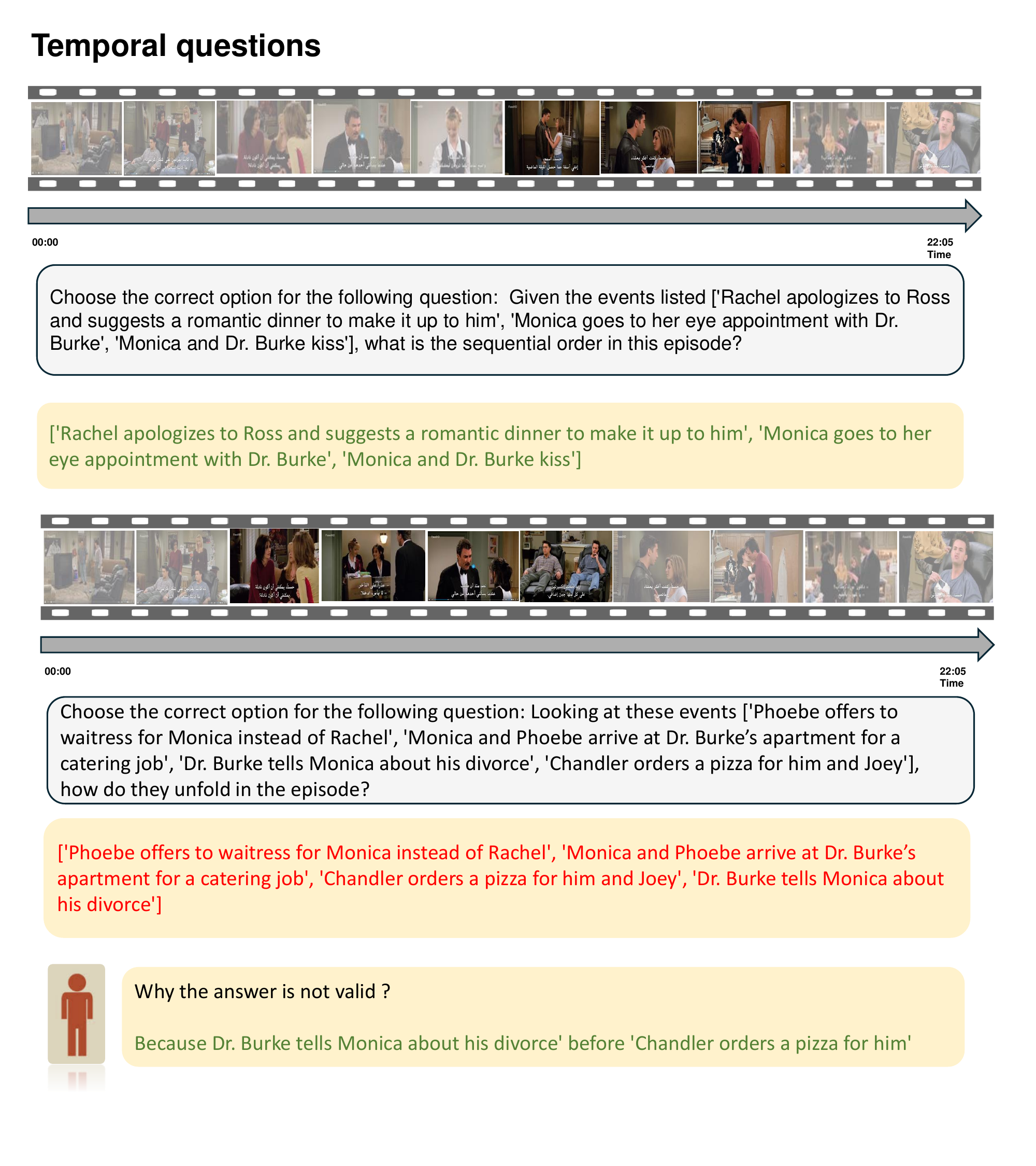}
    \caption{Examples of success and failure cases in Temporal Order of Events questions.}
    \label{temporal_failure_and_success}
\end{figure*}

In this section, we present examples of both success and failure cases in question generation using GPT-4o. Figure \ref{temporal_failure_and_success} illustrates cases involving the generation of chronological understanding questions, while Figure \ref{linking_failure_and_success} showcases examples related to Linking Multiple Events questions. As highlighted in the human evaluation section \ref{sec_appendix_human_verf}, such failure cases are infrequent, with 90.11\% of the generated data verified as accurate.

\subsection{Exact Prompts}
\label{sec:appendix_generation_deatils}
\begin{figure*}[ht]
    \centering
    \includegraphics[width=1\textwidth]{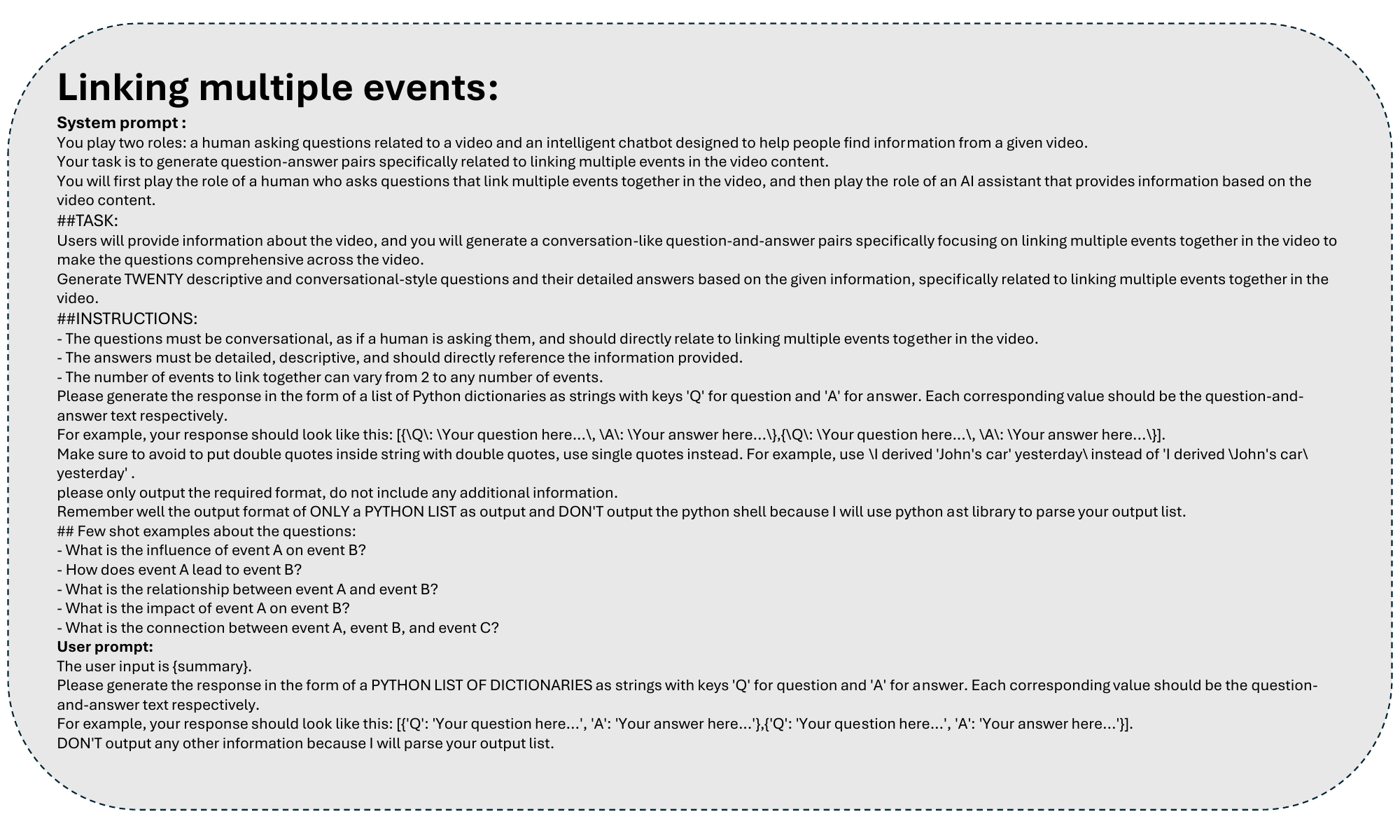}
    \caption{Detailed prompt for Linking multiple events questions generation.}
    \label{prompt_linking_events}
\end{figure*}

\begin{figure*}[ht]
    \centering
    \includegraphics[width=1\textwidth]{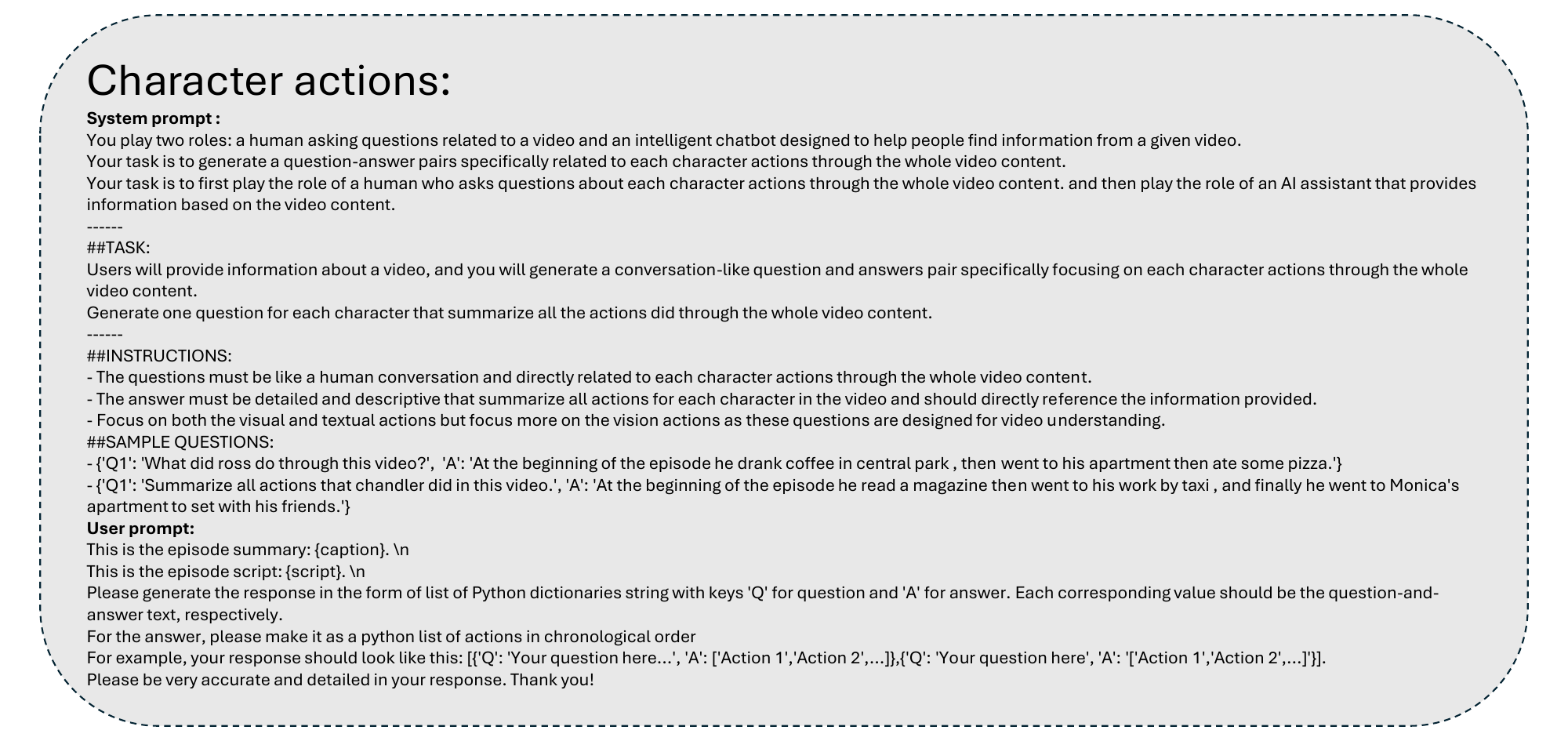}
    \caption{Detailed prompt for sequence of character actions questions generation.}
    \label{prompt_character_actions}
\end{figure*}

\begin{figure*}[ht]
    \centering
    \includegraphics[width=1\textwidth]{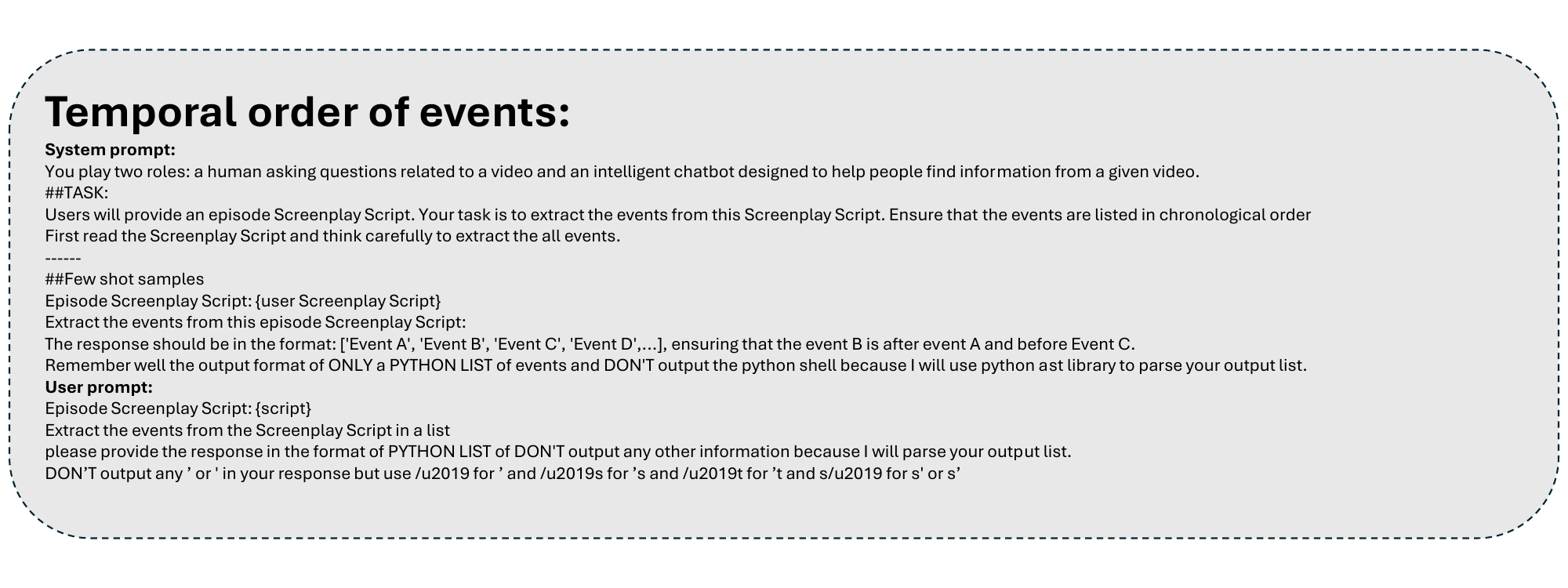}
    \caption{Detailed prompt for chronological understanding of events skill.}
    \label{prompt_temporal_events}
\end{figure*}

\begin{figure*}[ht]
    \centering
    \includegraphics[width=1\textwidth]{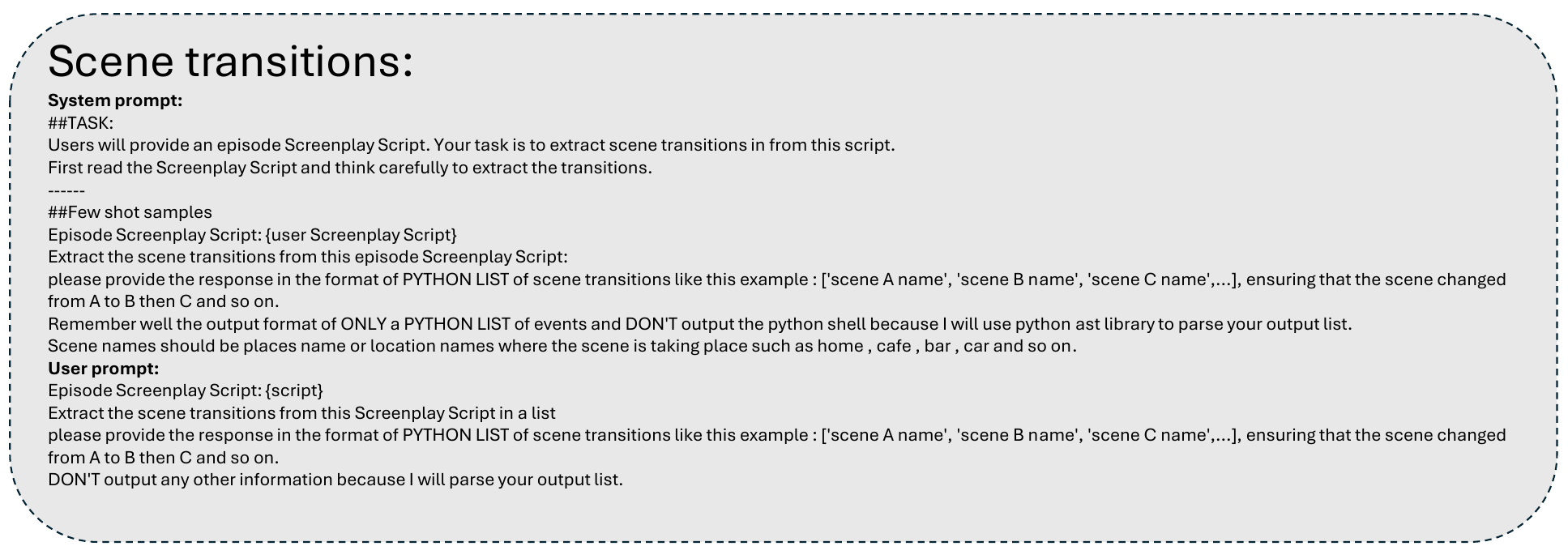}
    \caption{Detailed prompt for scene transitions questions generation.}
    \label{prompt_scene_transitions}
\end{figure*}

\begin{figure*}[ht]
    \centering
    \includegraphics[width=1\textwidth]{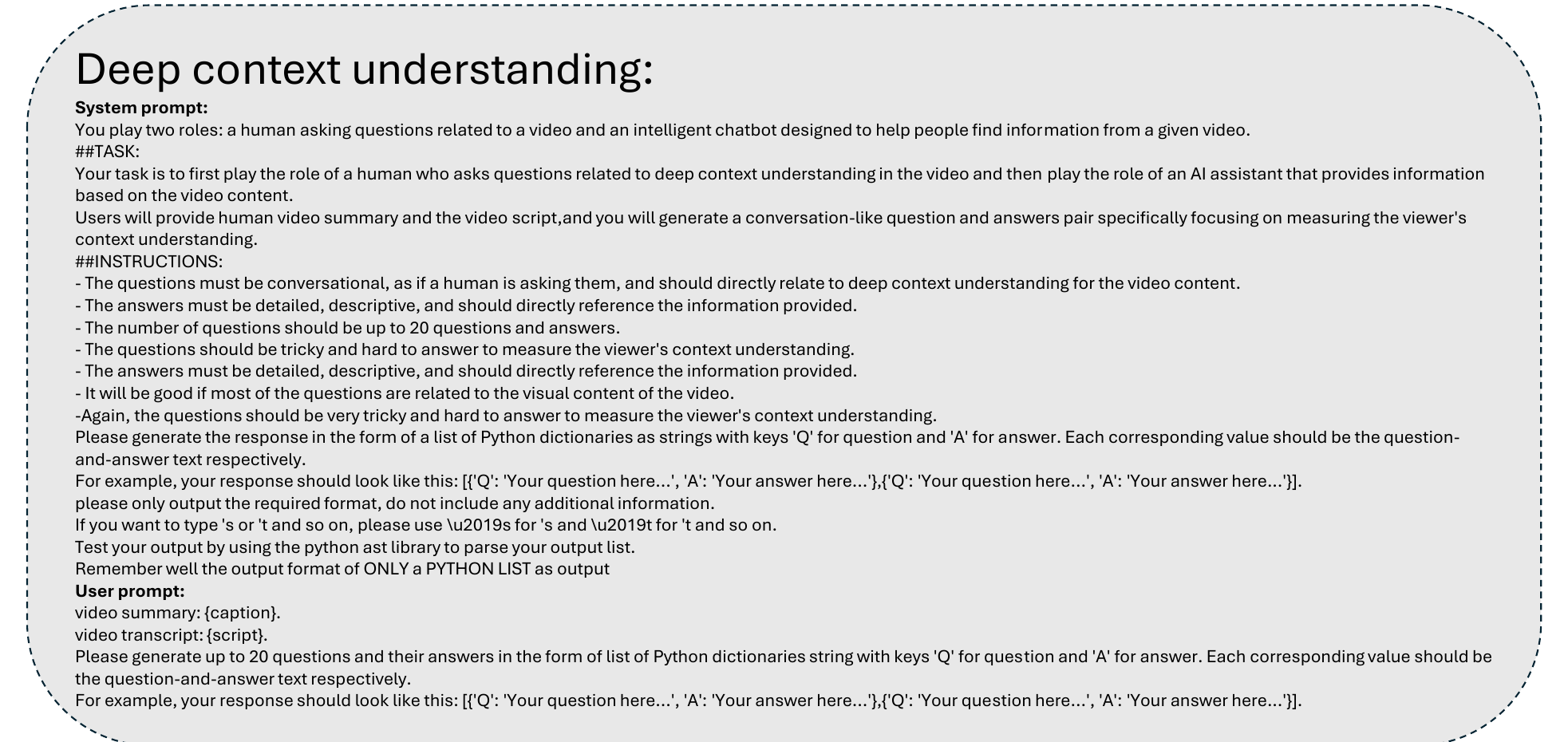}
    \caption{Detailed prompt for deep context understanding questions generation.}
    \label{prompt_context_unerstanding}
\end{figure*}

This section elaborates on the specific prompts employed to generate questions for each skill category. The prompts, utilized within the GPT-4o framework, are depicted in Figures \ref{prompt_linking_events}, \ref{prompt_temporal_events}, \ref{prompt_character_actions}, \ref{prompt_scene_transitions},\ref{prompt_context_unerstanding}.These figures provide the exact phrasing and structure used for question generation, ensuring reproducibility and clarity in the benchmarking creation process.

\section{Extended Related Work}
\label{sec:appendix_relatedwork}
\subsection{Semantic Roles in Vision and Language.}
Early work like \emph{Grounding Semantic Roles in Images}~\citep{silberer2018grounding} and \emph{Grounded Situation Recognition}~\citep{pratt2020grounded} showed how semantic roles can be aligned with visual regions in images. This concept was extended to videos by \emph{Video Object Grounding}~\citep{sadhu2020video} and \emph{Visual Semantic Role Labeling}~\citep{sadhu2021visual}, emphasizing semantic alignment for event understanding. Building on these ideas, HERO~\citep{li2020hero} proposed a hierarchical encoder that fuses visual and textual modalities, setting the stage for video-language pretraining. Recent models like VSGR~\citep{mao2022dynamic} further reinforce this trajectory by explicitly building video scene graphs to reason over semantic elements and their relations. In this context, our work introduces a benchmark that evaluates models' ability to track and reason about evolving semantic roles across long-form, multimodal narratives.
\subsection{Long Video Models.}
\label{Long_video_models}
Recent advancements in commercial AI models have introduced native support for long-form video understanding. For instance, Google's Gemini Flash family \citep{gemini_flash} that provides a one million-token context window for coherent multi-hour understanding, and GPT-4o \citep{GPT-4o} that processes video frames in conjunction with subtitles within its 128K-token context window. In contrast to the commercial solutions, Open-source efforts rely on compression, memory, and retrieval techniques. Qwen2.5-VL \citep{bai2025qwen25vltechnicalreport} extends dynamic resolution temporally to spot key events in hours-long streams and ground them visually. Qwen2-VL \citep{wang2024qwen2vlenhancingvisionlanguagemodels} uses Multimodal Rotary Position Embedding (M-RoPE) to encode up to 768 frames, preserving spatial and temporal structure. LongVU \citep{shen2024longvuspatiotemporaladaptivecompression} applies cross-modal queries and inter-frame dependencies for adaptive compression, trimming redundancy while keeping salient content. InternLM-XComposer-2.5-OmniLive \citep{zhang2024internlmxcomposer25versatilelargevision} pairs a short-term memory buffer with on-the-fly compression into long-term storage for later retrieval. Goldfish \citep{ataallah2024goldfishvisionlanguageunderstandingarbitrarily} segments videos into clips and retrieves the top-k most relevant, and LLaMA-VID \citep{li2023llamavid} represents each frame with only two tokens for maximal efficiency. VideoINSTA~\citep{liao2024videoinsta} enables long video understanding without pretraining by combining event-based temporal reasoning with LLMs and self-reflective spatial-temporal queries, achieving strong results on EgoSchema and NExT-QA. Despite these innovations, truly unrestricted long-form video understanding, with deep temporal dependencies and narrative complexity, remains an open challenge. 

\subsection{Video benchmarks.}

Recently, many multimodal large language models have aimed to extend their context lengths, such as Qwen2.5-VL~\citep{bai2025qwen25vltechnicalreport}, Qwen2-VL~\citep{wang2024qwen2vlenhancingvisionlanguagemodels}, and LongVU~\citep{wu2024longvideobench}, which can process videos over an hour long. This highlights the growing need for effective benchmarks. In contrast, short video benchmarks have been widely explored, covering tasks like moment retrieval~\citep{liang2024tvrrankingdatasetrankedvideo,lei2021qvhighlightsdetectingmomentshighlights}, action classification~\citep{caba2015activitynet,kinetics-400,soomro2012ucf101dataset101human}, video reasoning~\citep{xie2024funqasurprisingvideocomprehension,xiao2021nextqanextphasequestionansweringexplaining}, and video QA~\citep{xu2017video,lei2019tvqa,jang2017tgifqa}. Among longer video benchmarks, TVQA~\citep{lei2019tvqa} is an early example with over 152.5 K QA pairs across 21.8 K clips (1.27 min avg.), totaling 461.2 hours. More recent datasets like MovieChat-1K\citep{song2023moviechat} offer 1,000 movie clips (9.4 min avg.) with 14K annotations. CRAFT\citep{ates2020craft} and GazeVQA~\citep{ilaslan2023gazevqa} extend to causal and embodied reasoning. Other long-form efforts include MLVU~\citep{zhou2024mlvucomprehensivebenchmarkmultitask}, MoVQA~\citep{zhang2023movqa}, and Video-MME~\citep{videomme}, though their average durations peak at 17 minutes. LVBench\citep{wang2024lvbench} stands out with 1-hour videos, but includes only 103 clips totaling 117 hours. Domain-specific datasets like Ego4D QA on EgoSchema\citep{mangalam2023egoschema} focus on first-person views but use short (3-minute) clips. Motivated by these gaps, our benchmark evaluates grounding (temporal/visual anchoring) and reasoning (causality, narrative) skills to reflect human-like video understanding. As shown in Table~\ref{tab:benchmark_comparisons}, our dataset supports near-hour-long videos and offers the most significant total duration ($\sim$1K hours) and the largest QA-pair count in terms of long video understanding (\totalSampleNumber).


\begin{figure*}[ht]
    \centering
    \includegraphics[width=1\textwidth]{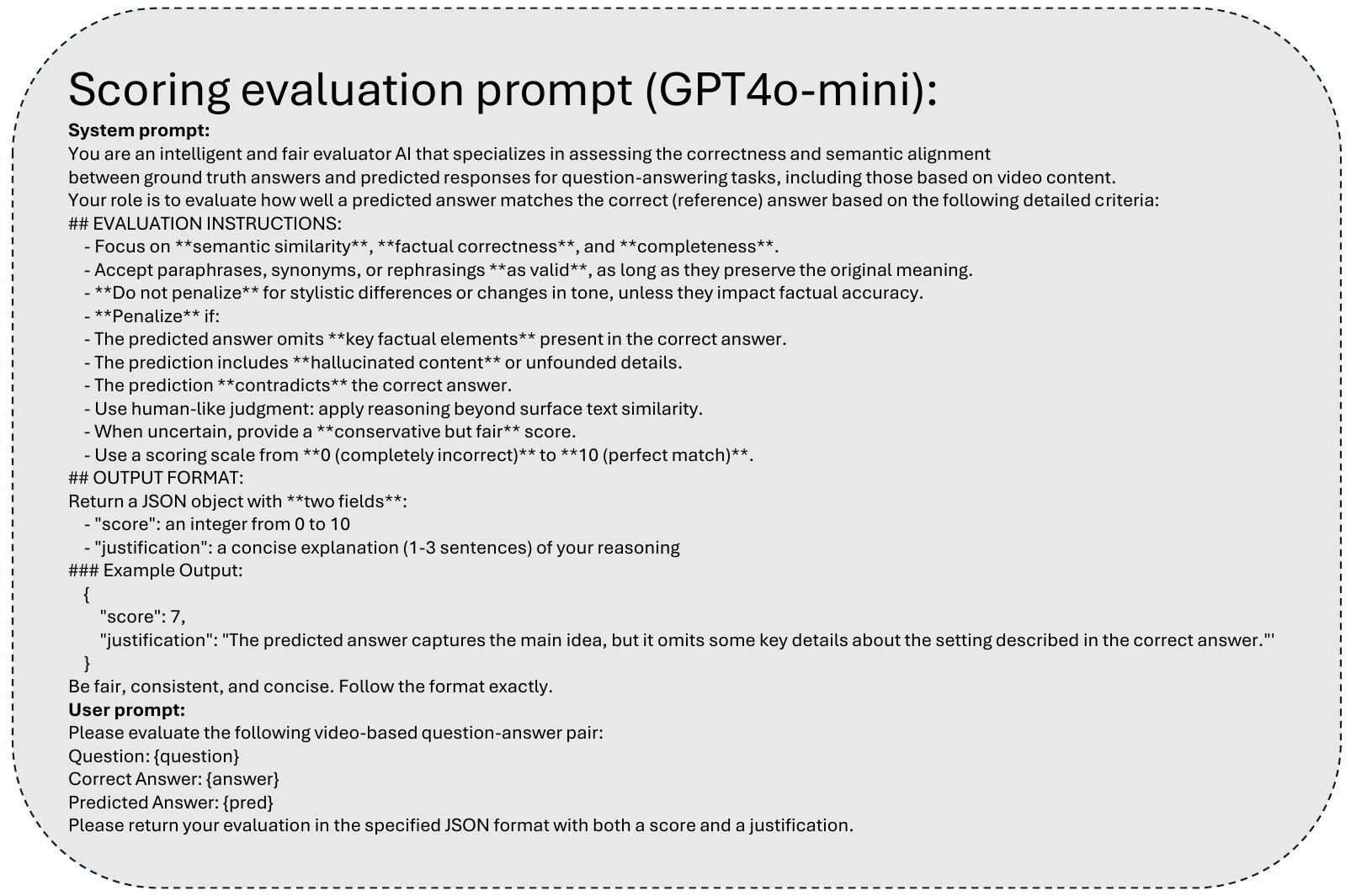}
    \caption{Detailed prompt for Scoring system evaluation.}
    \label{prompt_eval_score}
\end{figure*}


\label{sec:appendix_skills_examples}



\end{document}